\documentclass[11pt]{article}

\usepackage[preprint]{acl}
\usepackage{times}
\usepackage{latexsym}
\usepackage[T1]{fontenc}
\usepackage[utf8]{inputenc}
\usepackage{microtype}
\usepackage{inconsolata}
\usepackage{graphicx}

\usepackage{amsmath}
\usepackage{amssymb}
\usepackage{booktabs}
\usepackage{multirow}
\usepackage{colortbl}
\usepackage{tabularx}
\definecolor{OursAccent}{HTML}{008678}
\definecolor{OursTint}{HTML}{EAF5F2}
\usepackage[above,below]{placeins}
\usepackage{stfloats}
\graphicspath{{figures/}}

\title{Selective Reasoning for Typed Decision Models:\\When to Start, Read, and Stop}
\author{
  Bin Zheng\textsuperscript{$*$} \quad Zheng Chen\textsuperscript{$*$} \\
  {\normalfont University of Electronic Science and Technology of China} \\
  {\normalfont\texttt{zchen@uestc.edu.cn}}
}
\newcommand{\equalauthornote}{%
  \begingroup
  \renewcommand{\thefootnote}{\ensuremath{*}}%
  \footnotetext{Equal contribution.}%
  \endgroup
}

\begin{document}
\maketitle
\equalauthornote

\begin{abstract}
Typed decision models need an interface that can use additional reasoning while retaining candidate outputs. We train a shared candidate-answer reader on a frozen Qwen3.5-4B backbone to unify direct decisions with on-demand, interruptible reasoning. The reader returns distributions from empty and partial reasoning prefixes. A benefit gate starts reasoning, a natural-state trigger requests explicit reads, and an entropy rule accepts acquired answers and cancels generation. Measurements distinguish answer-query work, generated output, and pipeline time. Across 768 questions from ProofWriter, LogiQA~2.0, and ARC-Challenge, accuracy rises from 70.96\% for direct reading to 84.38\%. Relative to 512-token terminal reading, accuracy improves by 7.29 percentage points with 2.25\% more mean output work. Relative to 1,024-token terminal reading, pipeline time falls by 46.30\% with an accuracy difference of $-1.17$ points. Exact-prefix controls isolate the saving from interruption, while read-scheduling controls distinguish reduced observation work from shorter response time.

\end{abstract}

\section{Introduction}
\label{sec:introduction}
Typed decision models return a choice from supplied candidates, such as truth labels or answer indices, without generating an explanation. This interface suits repeated judgments within larger workflows. Kev implements it through a trained decision head on a language-model backbone \citep{palmer2026kev}. Explicit reasoning can construct inferences needed for some decisions \citep{wei2022chain}. We study how one candidate-answer interface can support immediate decisions, reasoning on demand, and accepted answers returned before reasoning finishes.

Reasoning can correct an initial answer or turn a correct decision into an error. Its expected benefit depends on both changes. Input-dependent budgets and selective-thinking methods adapt computation to the question \citep{snell2025scaling,zhang2025adaptthink,fang2025thinkless}. Once reasoning starts, intermediate answers offer opportunities to return early \citep{liu2025answer,boppana2026reasoning}.
% Generated plot: scripts/plot_frontpage_tradeoff.py; full-cohort primary estimates.
\begin{figure}[!t]
\centering
\includegraphics[width=\columnwidth]{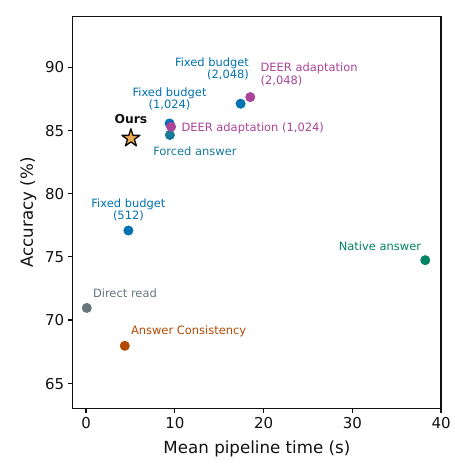}
\caption{Accuracy and mean pipeline time for ten configurations on the same 768 questions. The star marks Ours at $\tau=0.70$. Parenthesized numbers are reasoning-token caps; Fixed budget denotes terminal candidate reading.}
\label{fig:frontpage-tradeoff}
\end{figure}

Answer availability and answer acquisition are different operations. A natural generation state can indicate an opportunity to read, but obtaining a candidate distribution requires an explicit backbone query with an answer cue. Acceptance acts on this acquired distribution. Frequent reads incur input-processing work, whereas sparse reads can delay acceptance and consume more generation \citep{fu2025certaindex,xiang2026thinking,dong2026when}. Keeping the candidate reader fixed lets us measure these costs separately from changes to answer generation or parsing.

We train a shared candidate-answer reader on a frozen Qwen3.5-4B backbone \citep{qwen2026model} using empty and partial reasoning prefixes. A benefit gate estimates correction minus regression to select questions for reasoning. A natural-state trigger requests an explicit read, and an entropy rule accepts its answer and cancels active generation. The backbone retains its original generation policy until cancellation. DEER already separates transition monitoring, trial answers, and confidence-based exit \citep{yang2025dynamic}; our controller connects initial candidate decisions and partial-prefix reads through the same trained interface. Its trigger selects when to acquire a distribution rather than directly deciding when to exit.

Across 768 questions, selective reasoning improves candidate decisions at comparable output work and offers shorter pipeline time than 1,024-token terminal reading, as shown in Figure~\ref{fig:frontpage-tradeoff}. Separate matched controls verify interruption on identical token prefixes: acceptance and cancellation save 34.09\% of time with two fewer correct answers. Holding the gate fixed, learned reading reduces answer queries but increases time relative to fixed reads. These controls distinguish candidate-answer quality, observation work, and the time required to return a decision.

Our contributions are:
\begin{itemize}
  \setlength{\itemsep}{2pt}
  \setlength{\parsep}{0pt}
  \item[$\bullet$] \textbf{Unified candidate decisions.} One trained reader preserves the typed output interface across immediate decisions and decisions enhanced by interruptible reasoning.
  \item[$\bullet$] \textbf{Controlled execution evidence.} Matched token-prefix controls isolate the runtime saving from accepting an intermediate candidate answer and cancelling its continuing trajectory.
  \item[$\bullet$] \textbf{Acquisition trade-offs.} Fixed-reader comparisons expose how reading schedules exchange query and input-processing work for generation and response time.
\end{itemize}

\section{Related Work}
\label{sec:related}
\paragraph{Candidate decision interfaces.}
Kev returns supplied candidate selections through a trained decision head \citep{palmer2026kev}. Chain-of-thought prompting provides intermediate computation, and self-consistency aggregates sampled paths \citep{wei2022chain,wang2023selfconsistency}. We train one reader across empty and partial prefixes on frozen Qwen3.5-4B \citep{qwen2026model}, retaining the same candidate outputs for direct and reasoning-enhanced decisions.

\paragraph{Reasoning initiation and budgets.}
AdaptThink, Thinkless, and AutoThink select direct or reasoning-assisted responses \citep{zhang2025adaptthink,fang2025thinkless,tu2025autothink}. Sonata uses prefill states to predict the fraction of correct sampled non-thinking answers, termed self-consistency in that work, and extends this prediction to thinking gain \citep{li2026adaptive}. Compute allocation, budget forcing, and shorter-chain selection control how much reasoning is used \citep{snell2025scaling,muennighoff2025s1,hassid2025dontoverthink}. Our gate estimates corrections minus regressions from paired outcomes of the shared candidate reader. Initial routing connects to that reader's intermediate acceptance and cancellation policy.

\paragraph{Intermediate answers and early stopping.}
Answer Convergence and Dynasor assess trial-answer consistency or certainty \citep{liu2025answer,fu2025certaindex}. DEER monitors transitions, induces trial answers, and accepts confident answers, with branch-parallel decoding to overlap trials and continued reasoning \citep{yang2025dynamic}. DTSR screens reflection signals before requesting a sufficiency score \citep{xiang2026thinking}. Our reader acquires a fixed-candidate distribution that a separate rule tests for acceptance. The same reader supplies initial decisions and intermediate answers. Reads use separate engine requests, whereas DEER's branch-parallel implementation uses a specialized attention mask and dynamic KV management.

\paragraph{Internal signals and execution cost.}
Reasoning Theater probes answer availability \citep{boppana2026reasoning}; NEAT and BLADE use internal signals to control exit or reflection \citep{liu2026neat,fu2026blade}. LYNX trains on forced-exit outcomes and applies a hidden-state probe at natural cues to decide exit \citep{akgul2025lynx}. LearnStop predicts stopping from probed answers and studies query density and cache costs \citep{dong2026when}. Our natural-state trigger selects acquisition before a candidate distribution is available. Acceptance acts on the subsequent read. We hold the candidate reader fixed and measure online acquisition and cancellation, with exact-prefix controls isolating interruption from reasoning trajectory changes.
\par
\begin{figure*}[t]
\centering
\includegraphics[width=\textwidth]{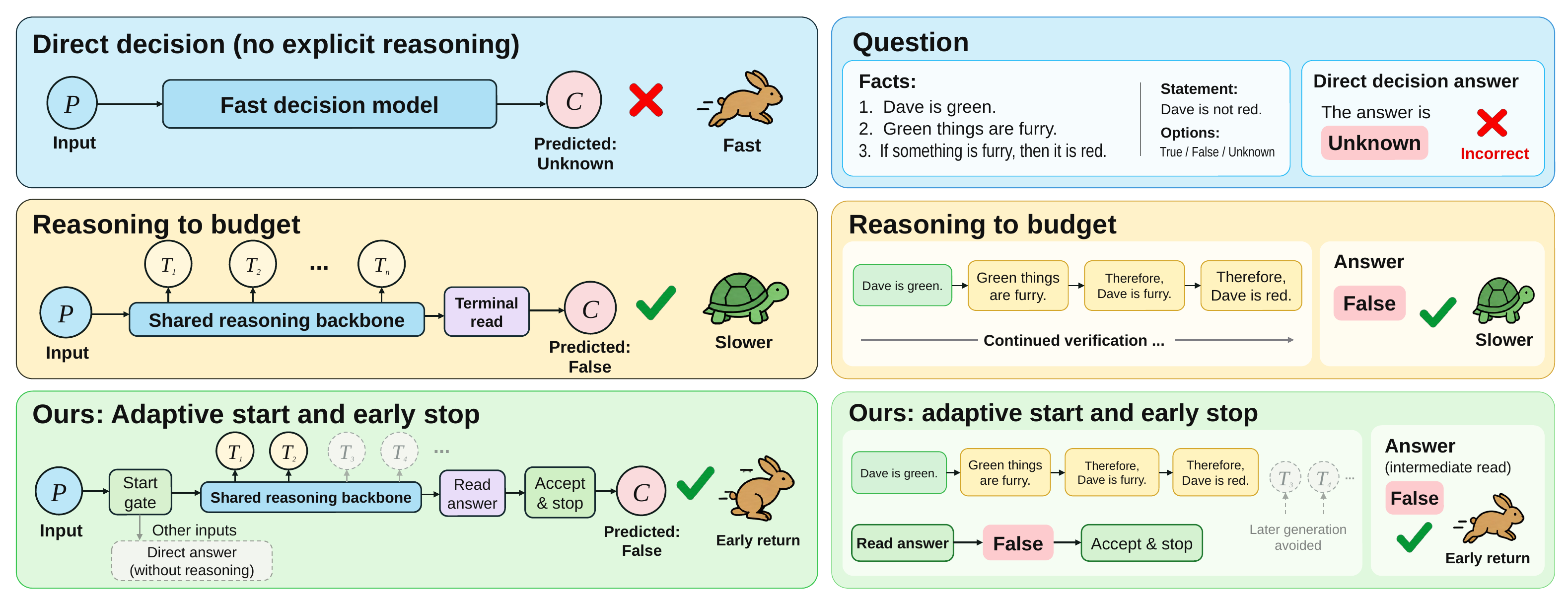}
\caption{On-demand, interruptible reasoning through a shared candidate interface. The abridged ProofWriter example shows reasoning correcting a direct answer and intermediate acceptance avoiding later generation. The start gate also permits direct return. $P$, $T_i$, and $C$ denote the prompt, thought fragments, and candidate answer. Grey thoughts mark avoided generation.}
\label{fig:method}
\end{figure*}

\section{Unified Candidate Decisions with Interruptible Reasoning}
\label{sec:method}
A shared candidate-answer reader supports direct decisions from an empty reasoning prefix and enhanced decisions from partial reasoning. The decision process can return immediately, start reasoning on demand, or return an accepted intermediate answer and cancel unfinished generation. Both paths retain the same candidate output interface on a frozen backbone.

The framework separates three decisions: whether to initiate reasoning, when to acquire an answer, and whether to accept it. Our implementation uses a benefit gate, a trigger based on natural generation states and prefix length, and entropy-based acceptance. Figure~\ref{fig:method} summarizes these decisions and the direct-return alternative. The reader and two learned control modules are fitted separately; backbone weights remain frozen.

\subsection{A Shared Candidate-Answer Reader}
Input $x$ contains a question, accompanying material, and $K$ candidate answers. A frozen Qwen3.5-4B backbone~\citep{qwen2026model} supports autoregressive reasoning and an explicit answer reader. Let $z_{1:t}$ be a generated prefix, with $z_{1:0}$ empty. The reader uses the empty prefix for direct decisions and a generated prefix for reasoning-enhanced decisions. A read processes $x$ and this prefix, appending \texttt{</think>} if the prefix lacks a thinking closure, followed by a \texttt{<FINAL>} answer cue. A bias-free linear head maps the state $\tilde h_t$ at the cue's last token to candidate probabilities:
\begin{equation}
 p_t=\operatorname{softmax}(W_{1:K}\tilde h_t),\quad
 a_t=\arg\max_{j\leq K}p_t(j).
 \label{eq:reader}
\end{equation}
Here $W\in\mathbb{R}^{4\times2560}$ has 10,240 parameters, and Eq.~\ref{eq:reader} applies when $K\leq4$. For more candidates, we normalize the native language-model logits over candidate-index tokens at the same position. Both interfaces require the backbone to process the answer query. Each query emits one auxiliary native token to complete the request, while the prediction is obtained from $p_t$.

An initial read supplies $a_0$, $p_0$, and $\tilde h_0$. The controller either returns $a_0$ or starts reasoning up to budget $B$. The natural state $h_t$ is the final backbone state passed to the vocabulary projection at the last token of $z_{1:t}$. It is saved from decoding and indexed by prefix length. The trigger uses this state when retained; obtaining $\tilde h_t$ requires an explicit answer query. Retention and availability follow the state interface in Appendix~\ref{sec:controller-interface}.

Training fits the reader across both prefix types while keeping the backbone frozen. It minimizes
\begin{equation}
 \mathcal{L}_{\mathrm{read}}
 =-\mathbb{E}_{i,t}\left[\log p_{i,t}(y_i)\right],
 \label{eq:reader-loss}
\end{equation}
where the expectation is uniform over training questions $i$ and checkpoints $t\in\mathcal{T}=\{0,128,256,512,1024\}$. Positions are capped at each trace's available length. The selected head is fixed for online evaluation; optimization details appear in Appendix~\ref{sec:hyperparameters}.

\subsection{When to Start: Expected Corrective Benefit}
The initiation decision selects direct return or a reasoning trajectory. Our implementation estimates corrective benefit from paired initial and terminal answers under cap $B$, using the same reader. For training supervision, let $c_0=\mathbb{I}[a_0=y]$ and $c_B=\mathbb{I}[a_B=y]$ denote their correctness. Correctness can increase by one, decrease by one, or remain unchanged. Thus,
\begin{equation}
\begin{aligned}
 \mathbb{E}[c_B-c_0\mid\phi_0]
 &=\Pr(c_0=0,c_B=1\mid\phi_0)\\
 &\quad-\Pr(c_0=1,c_B=0\mid\phi_0).
\end{aligned}
\label{eq:benefit}
\end{equation}
We estimate the two events with logistic models:
\begin{align}
 g(x)&=\sigma(w_g^\top\phi_0+b_g),\\
 d(x)&=\sigma(w_d^\top\phi_0+b_d).
\end{align}
The score $u(x)=g(x)-d(x)$ estimates corrective benefit. Input $\phi_0$ concatenates the initial answer state with maximum probability, top-two margin, normalized entropy $H(p_0)/\log K$, and $\log(1+|x|)$, where $|x|$ is the input token count. Feature standardization is fitted on gate-training groups and folded into deployed weights. The controller starts when $u(x)>\theta$; otherwise it returns $a_0$.

Each event model minimizes L2-regularized binary cross-entropy using out-of-fold reader predictions from training material groups. The deployed score uses the raw logistic probabilities. A separate threshold split sets $\theta$ for a target reasoning fraction of 0.75. Actual routing rates and token costs can vary on new questions. The gate targets the $B=1024$ terminal endpoint, and its combination with early stopping is evaluated as a complete policy. Appendix~\ref{sec:reproducibility} specifies regularization and fitting settings.

\subsection{When to Read: Selective Answer Queries}
The acquisition decision schedules explicit candidate-answer queries. Our implementation screens prefixes using the natural state $h_t$ and prefix length before paying for a read. Its deployed computation is
\begin{equation}
 r_t=\sigma(w^\top h_t+\alpha t/1024+b).
 \label{eq:trigger}
\end{equation}
The trigger uses no task identity. At inference, $p_t$ is computed only after a read is requested: $r_t$ predicts a reading opportunity, and the read supplies the answer and certainty.

Reading opportunities come from complete-sentence boundaries. With generation active and the reader idle, the trigger considers the latest pending prefix when its absolute sentence ordinal satisfies $s\bmod4=0$ and its indexed $h_t$ is available. Each scored event is consumed once, requesting a read if $r_t\geq0.25$. New sentence events replace the pending prefix, so coalescing can skip checkpoints. A submitted read remains bound to its selected token prefix while generation continues. Boundaries align to original token IDs and exclude the next sentence's content. Appendix~\ref{sec:controller-interface} gives the waiting, replacement, and completion rules.

Training supervision identifies prefixes with correct, concentrated reader predictions. Define entropy certainty by
\begin{equation}
 q_t=1-\frac{H(p_t)}{\log K}.
\label{eq:certainty}
\end{equation}
At training time, an out-of-fold reader supplies
\begin{equation}
 v_{i,t}=\mathbb{I}[a_{i,t}=y_i\ \land\ q_{i,t}\geq0.85].
\end{equation}
A logistic trigger minimizes L2-regularized weighted binary cross-entropy for $v_{i,t}$, assigning equal total weight to each question. Its training features are the natural state projected onto 64 training-fitted principal components and $t/1024$. Standardization and projection fold into Eq.~\ref{eq:trigger}; Appendix~\ref{sec:reproducibility} gives the fitting configuration.

\subsection{When to Stop: Accepting an Observed Answer}
Acceptance acts on an acquired candidate distribution. Our implementation accepts the prefix-bound answer $a_t$ when its entropy certainty satisfies $q_t\geq\tau$ and cancels generation if it is active. The representative policy uses $\tau=0.70$; $0.85$ and $0.95$ provide additional operating points with the same gate and trigger. The trigger's training target stays at 0.85. Entropy certainty measures concentration of the answer distribution; its threshold does not specify a probability of correctness.

A rejected read leaves generation active and allows the latest pending event to be considered. At natural completion or the budget limit, the controller finishes any outstanding read. If no answer is accepted, a terminal read supplies the answer without a certainty requirement. A prefix answer accepted after generation ends is recorded separately from cancellation that saves generation.

\subsection{Online Execution and Cost}
Generation and answer requests share one backbone and one vLLM engine \citep{kwon2023pagedattention}, with at most one answer query in flight. Starting reasoning submits a separate request on $x$. It remains scheduled while a read processes a separate prefix plus answer cue; the cue never enters the reasoning trace. Overlap shares GPU resources and can affect generation throughput. The configuration retains autoregressive KV state within a request and disables cross-request prefix caching, so generation and each answer query incur their own input-processing work.

For the complete controller, answer requests count one initial read, all intermediate reads, and a terminal read if needed. Trigger calls count actual evaluations of $r_t$. Output tokens include natural generation and one auxiliary token per read. The input-token counts in Table~\ref{tab:controller-work} sum the generation prompt and every answer-query input.

For an accepted read, let $t$ be the queried prefix length, $d$ the generated length when the decision arrives, and $c$ the length when cancellation takes effect. We record $c-t$ and $c-d$ as subsets of the natural output already counted. Pipeline time spans initial reading, routing, generation, sentence detection, state capture, triggers, answer reading, and cancellation under the event-driven procedure in Table~\ref{alg:controller}. Component durations describe overlapping work and are not summed into a latency estimate.

\setcounter{dbltopnumber}{1}
\section{Experimental Setup}
\label{sec:experiments}
\subsection{Tasks and Model Fitting}
\paragraph{Evaluation groups.}
We evaluate 256 questions each from ProofWriter, LogiQA~2.0, and ARC-Challenge, covering open-world logical judgments, passage-based reasoning, and scientific knowledge questions \citep{tafjord2021proofwriter,liu2023logiqa,clark2018arc}. ProofWriter and LogiQA contribute one question per reserved official-training material group, disjoint from fitting, development, and earlier component evaluation. Groups connect shared source identities and identical or near-duplicate material. ARC uses an official-test subset with development overlap excluded. Candidate order is preserved. ProofWriter and LogiQA require material-grounded answers, while ARC permits scientific knowledge and tests transfer of the fitted components.

\paragraph{Fitting and selection.}
Fitting uses 1,024 training and 256 development questions from ProofWriter and LogiQA under the data-isolation protocol in Appendix~\ref{sec:reproducibility}. Reader selection minimizes development negative log-likelihood on a frozen backbone. Gate fitting and threshold selection use separate training-material partitions. Out-of-fold reader predictions supervise the gates and trigger, with equal trigger-training weight per question. The reader, control parameters, and three acceptance thresholds are fixed before primary evaluation.

\subsection{Comparison Policies}
Our comparisons address three questions. Shared-reader policies test computation allocation under a fixed candidate interface. Generated-answer baselines compare complete configurations with different answer and stopping interfaces. Matched controls explain the roles of initiation, reading, and intermediate acceptance within our controller.

Eight shared-reader configurations compare fixed budgets with selective allocation using identical backbone and reader weights. Direct reading in Table~\ref{tab:main-results} establishes initial accuracy, and terminal reading at caps of 512, 1,024, and 2,048 tokens measures reasoning gains. The simple controller combines margin routing, reading after at least eight additional sentences, and entropy acceptance at 0.85. Our controller uses acceptance thresholds 0.70, 0.85, and 0.95, with their operating points reported in Table~\ref{tab:threshold-operating-points}. Both start thresholds are calibrated to a target reasoning fraction of 0.75.
% Generated by scripts/render_results.py; edit the renderer, not the cells.
\begin{table*}[t]
\centering\small
\setlength{\tabcolsep}{3.5pt}
\begin{tabularx}{\textwidth}{@{}>{\raggedright\arraybackslash}Xrrrrrrrr@{}}
\toprule
Method & Cap & PW (\%) & LQ (\%) & ARC (\%) & Avg. (\%) & Time (s) & Output tok. & Queries\\
\midrule
\multicolumn{9}{l}{\textit{Shared-reader baselines}}\\
Direct read & 0 & 58.20 & 66.41 & 88.28 & 70.96 & 0.080 & 1.0 & 1.00\\
Terminal read & 512 & 67.97 & 68.75 & 94.53 & 77.08 & 4.773 & 513.0 & 1.00\\
Terminal read & 1,024 & 83.59 & 76.95 & 96.09 & 85.55 & 9.410 & 1,022.7 & 1.00\\
Terminal read & 2,048 & 85.55 & 78.91 & 96.88 & 87.11 & 17.414 & 1,887.7 & 1.00\\
\midrule
\multicolumn{9}{l}{\textit{Generated-answer baselines}}\\
Forced terminal answer & 1,024 & 80.86 & 76.95 & 96.09 & 84.64 & 9.446 & 1,025.1 & 1.00\\
Answer Consistency ($k=10$) & 1,024 & 41.02 & 71.88 & 91.02 & 67.97 & 4.362 & 307.0 & 12.28\\
DEER adaptation & 1,024 & 80.86 & 78.91 & 96.09 & 85.29 & 9.582 & 1,026.7 & 1.76\\
DEER adaptation & 2,048 & 86.33 & 80.47 & 96.09 & 87.63 & 18.500 & 1,902.7 & 4.43\\
Native answer & 8,192 & 79.69 & 60.16 & 84.38 & 74.74 & 38.213 & 4,141.7 & 0.00\\
\midrule
Ours ($\tau=0.70$) & 1,024 & 83.98 & 75.39 & 93.75 & 84.38 & 5.053 & 524.5 & 2.43\\
\bottomrule
\end{tabularx}
\caption{Accuracy and per-question costs on 768 questions, 256 per task. PW, LQ, and ARC denote ProofWriter, LogiQA~2.0, and ARC-Challenge; Avg. is task-mean accuracy. Cap bounds reasoning tokens, including the native final answer. Output includes auxiliary tokens; Queries counts answer/trial requests. Time is mean pipeline seconds.}
\label{tab:main-results}
\end{table*}

Generated-answer policies compare alternative answer and stopping interfaces using the settings in Appendix~\ref{sec:baseline-settings}: forced terminal answering at 1,024 tokens, Answer Consistency (AC), a candidate-answer adaptation of Answer Convergence using ten consecutive matching trials \citep{liu2025answer}, Qwen-adapted DEER at 1,024 and 2,048 tokens \citep{yang2025dynamic}, and native generation at 8,192 tokens. All start reasoning, with natural completion allowed before the cap. Trial and forced-final outputs count additionally, while the native cap includes the final answer.

The supplementary comparisons in Appendix~\ref{sec:calibrated-policies} select simple reading intervals and DEER trial thresholds on 48 development questions disjoint from evaluation, then compare controllers at corresponding acceptance thresholds. These development questions select policies; the separate 48-question timing subset comes from evaluation and plays no role in selection.

\subsection{Execution and Measurement}
All configurations use Qwen3.5-4B on A100 80GB with BF16 vLLM, greedy decoding, compiled execution, and disabled cross-request prefix caching.

We report per-task and task-equal accuracy, and mean pipeline latency. Invalid generated answers count as errors. Synchronized timing covers initial reading, routing, generation, sentence detection, state capture and synchronization, triggers, queries, and cancellation. Loading, warm-up, tokenization, cache resets, and record writes are outside the timer. Input and output tokens, answer/trial queries, and trigger calls characterize computational work. Outputs include auxiliary tokens and generation beyond an accepted prefix.
% Queue the full-width results plot one page before its substantive analysis.
% Generated by scripts/render_results.py; vector plot: scripts/plot_accuracy_cost.py.
\begin{figure*}[t]
\centering
\includegraphics[width=\textwidth]{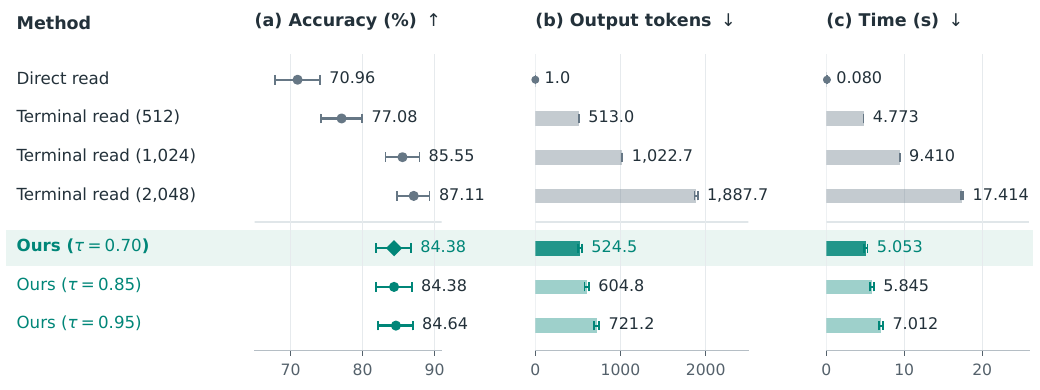}
\caption{Accuracy and cost for seven shared-reader configurations on the same 768 questions and A100. Output work and pipeline time include auxiliary queries. Whiskers are pointwise 95\% material-group bootstrap intervals. Terminal labels give reasoning caps; shading marks the predefined $\tau=0.70$ policy.}
\label{fig:accuracy-cost}
\end{figure*}

Principal estimates use first executions on the same 768 questions for all 13 configurations. Figure~\ref{fig:accuracy-cost} plots accuracy and both costs for seven shared-reader configurations. Matched controls interleave policy executions within each question. Table~\ref{tab:same-device-timing} reports first-execution timing on 48 evaluation questions, 16 per task. Two additional executions support repeated timing analyses, averaging within question. Pointwise 95\% intervals use 10,000 paired material-group bootstrap resamples, stratified by ProofWriter subtype, LogiQA, and ARC, as specified in Appendix~\ref{sec:cost-details}.

\subsection{Matched Mechanism Analysis}
We compare \textit{Ours} with three component removals and two policy replacements on the same 768 questions. All share the reader and 1,024-token cap, with $\tau=0.70$ for intermediate acceptance when enabled. Policies interleave within each question against a local complete-controller execution. Table~\ref{tab:ablations} reports the removals; Table~\ref{tab:control-work} includes both removals and replacements with complete work metrics.

\textit{Always reason} forces generation after retaining the initial read and gate computation, testing selective initiation. \textit{No early stop} retains intermediate queries but disables intermediate acceptance and cancellation, testing interruptible return. \textit{Gate + terminal read} jointly removes intermediate reads and early acceptance, testing the full process policy against routing followed by terminal reading.

\textit{Margin gate} replaces benefit scoring with candidate-probability margin routing, while retaining the downstream reading and acceptance policies. It tests corrective-benefit modeling against uncertainty-based allocation. \textit{Fixed reads} requests an answer after at least eight additional sentences, while retaining the start gate, asynchronous latest-prefix handling, and acceptance rule. It tests how a reading schedule exchanges query overhead for answer-acceptance timing.

Routing diagnostics compare initial answers with the deployed \textit{Always reason} policy's outcomes, whereas gate training targets terminal-1,024 correctness. Stopping analysis checks exact token-prefix agreement and distinguishes cancellation from acceptance after generation ends.

Supplementary studies in Appendix~\ref{sec:supplementary-comparisons} cross both gates and reading schedules and compare serial with concurrent execution, reusing the observed evaluation cohort and original controller.

\section{Results and Analysis}
\label{sec:analysis}
\subsection{Unified Direct and Reasoning-Enhanced Decisions}
\label{sec:main-results}
Optional reasoning strengthens decisions through a unified candidate-answer interface. Across 768 questions, the predefined $\tau=0.70$ controller raises accuracy from 70.96\% for direct reading to 84.38\%: 648 rather than 545 correct answers, or 103 additional correct decisions and a gain of 13.41 percentage points, as reported in Table~\ref{tab:main-results} and Figure~\ref{fig:accuracy-cost}. The same candidate-answer reader supports immediate decisions and answers acquired from partial reasoning. Paired 95\% confidence intervals accompany the primary and matched contrasts in Table~\ref{tab:paired-comparisons}.

Selective allocation also obtains more correct answers at comparable average output work. The 512-token terminal reader answers 592 questions correctly, reaching 77.08\% accuracy, while the controller adds 56 correct answers, improving accuracy by 7.29 percentage points. Mean output rises from 513.0 to 524.5 tokens, or 2.25\%, and mean pipeline time from 4.773 to 5.053 seconds, or 5.87\%.

Relative to 1,024-token terminal reading, the controller reduces pipeline time by 46.30\% and output work by 48.71\%. Accuracy changes from 85.55\% to 84.38\%, a paired difference of $-1.17$ points.

The predefined simple controller in Appendix~\ref{sec:calibrated-policies} reaches 85.16\% at 5.834 seconds; its development-calibrated variant reaches 648/768 at $\tau=0.70$. DEER reaches 85.29\% at 1,024 tokens and 87.63\% at 2,048.

\subsection{Interruptible Reasoning through Intermediate Answer Acceptance}
\label{sec:interruptible-reasoning}
% Generated by scripts/render_results.py; edit the renderer, not the cells.
\begin{table}[t]
\centering\small
\setlength{\tabcolsep}{3.5pt}
\begin{tabular*}{\columnwidth}{@{\extracolsep{\fill}}lrrr@{}}
\toprule
Variant & Acc. (\%) & $\Delta$Acc. (pp) & $\Delta$Time (\%)\\
\midrule
Ours & 84.90 & -- & --\\
\midrule
No early stop & 85.16 & $+0.26$ & $+51.71$\\
Gate + terminal read & 83.98 & $-0.91$ & $+43.30$\\
Always reason & 85.55 & $+0.65$ & $+23.04$\\
\bottomrule
\end{tabular*}
\caption{Component removals on 768 questions with cap 1,024 and $\tau=0.70$. Matched Ours answers 652 correctly in 4.960 s. Deltas subtract Ours; time changes are normalized by its mean. Gate + terminal read removes intermediate queries and early acceptance. Table~\ref{tab:control-work} reports complete costs and policy replacements.}
\label{tab:ablations}
\end{table}

Holding the start gate and reading trigger fixed, intermediate answer acceptance and cancellation reduce mean time from 7.524 to 4.960 seconds, saving 34.09\%, while correct answers change from 654 to 652. The controls in Table~\ref{tab:ablations} use a separate matched execution on the same 768 questions, with 652 correct answers for the complete-controller reference, compared with 648 in the primary execution in Table~\ref{tab:main-results}.

\begin{figure}[t]
\centering
\includegraphics[width=\columnwidth]{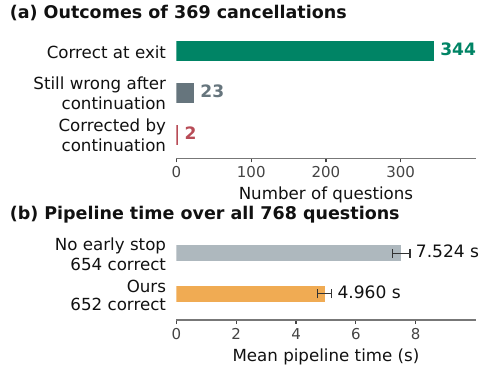}
\caption{Stopping outcomes and pipeline time against the matched no-early-stop control. Panel (a) compares 369 cancelled answers with continuation to natural completion or 1,024 tokens. Panel (b) averages all 768 questions. Whiskers show pointwise 95\% paired material-group bootstrap intervals.}
\label{fig:stopping-mechanism}
\end{figure}

Exact-prefix controls support attributing this saving to interrupting the reasoning trajectory. All 369 cancelled traces are exact token prefixes of the no-stop traces, and the other 399 traces are identical. Initial answers and start decisions agree on every question. Figure~\ref{fig:stopping-mechanism} separates wrong exits from errors recoverable by continuation: of 25 wrong exits, 23 remain wrong and two are corrected. Those two recovered answers represent 0.26 percentage points over all 768 questions.

Further reasoning can correct a wrong intermediate answer along the same trajectory. The paired control measures these forgone corrections alongside the computation avoided.
% Vector figure rendered by scripts/plot_stopping_mechanism.py --publish.

Against routing followed by terminal reading, the controller answers seven more questions correctly and saves 30.22\% of time. This comparison jointly measures intermediate acquisition and early acceptance.

\subsection{Selective Initiation of Reasoning}
\label{sec:selective-initiation}
Selective starting allows the framework to return direct decisions without launching generation on 191 questions. Compared with \textit{Always reason}, it reduces reasoning invocations from 768 to 577 and saves 18.72\% of pipeline time, with correct answers changing from 657 to 652. This allocates additional computation to a subset of questions while retaining the direct-answer path.

The choice of start score determines which correction and regression opportunities receive reasoning. The routing diagnostics in Appendix~\ref{sec:mechanism-details} show that benefit routing avoids more regressions but selects fewer corrections than margin routing. Benefit-minus-margin accuracy differs by $-0.39$ points, and time by $+0.121$ seconds. Neither these contrasts nor the factorial controls in Appendix~\ref{sec:factorial-controls} establish an advantage for benefit over margin.

\subsection{Read Scheduling and Execution Cost}
\label{sec:read-scheduling}
Reading less often reduces observation work while changing when an answer can be accepted. Under the same gate and stopping rule, learned reading reduces queries from 4.105 to 2.427 per question, or 40.88\%. Correct answers change from 643 to 652, a difference of $+1.17$ points. Mean time rises from 4.554 to 4.960 seconds, a paired increase of $0.406$ seconds. Fewer reads can defer acceptance and allow more generation.

An intermediate read processes a fixed prefix while generation can continue on the same engine. Its response delay can allow generation to extend beyond that prefix, linking query frequency to observation workload and return time.

The calibrated simple policy provides a second view of this trade-off. At $\tau=0.70$, both complete controllers answer 648/768 correctly, while the learned controller makes 42.75\% fewer queries. Repeated timing in Appendix~\ref{sec:calibrated-policies} resolves no latency advantage. This separates reduced observation work from the time required to return an answer.

At the trigger's original training target, the cached development diagnostic in Appendix~\ref{sec:position-diagnostic} finds better pooled answer-availability prediction from natural states, without improved within-question temporal ranking. Follow-up controls examine trigger-target alignment and bounded waiting in Appendix~\ref{sec:trigger-revisions}, and serial versus concurrent reading in Appendix~\ref{sec:execution-comparison}. Read scheduling must balance query costs with the generation consumed before acceptance: query count alone does not determine response latency.

\section{Conclusion}
One shared candidate-answer reader unifies direct decisions with on-demand, interruptible reasoning on a frozen backbone. Empty and partial reasoning prefixes retain the same typed output interface, allowing the controller to allocate computation without changing how answers are returned. Across three tasks, this unified process improves accuracy at comparable output work and reduces time relative to longer terminal reading.

Matched token-prefix controls isolate the saving from accepting an intermediate answer and cancelling unfinished generation, with the preceding trajectory held fixed. Read-scheduling controls show that fewer queries reduce observation work but can delay acceptance and consume more generation. Efficient candidate decisions therefore require joint control over initiation, answer acquisition, and acceptance, with query costs and computation before return measured together.

\clearpage
\setcounter{dbltopnumber}{2}
\section*{Limitations}
We evaluate one frozen Qwen3.5-4B backbone, fitting on ProofWriter and LogiQA and testing transfer on ARC-Challenge. Material groups separate fitting and evaluation. Controller supervision differs from online acceptance, and entropy certainty measures answer concentration.

Our measurements cover single-question pipeline time and token work. Energy and service throughput remain unmeasured. Separate greedy runs can produce different trajectories, so comparisons retain local paired executions. Follow-up policies reuse observed questions; choosing a revised controller requires evaluation on fresh data.

\section*{Ethics Statement}
This work uses public reasoning benchmarks and model inference, without a user study or deployment in consequential decisions. Early-stopping errors can propagate through a downstream workflow; applications with unequal error costs need task-specific acceptance criteria. Dataset and model use and any redistribution must follow the original licenses. We report accuracy alongside computational cost so that efficiency gains do not obscure changes in answer quality.

\bibliography{references}

\begin{thebibliography}{24}
\providecommand{\natexlab}[1]{#1}

\bibitem[{Akg{\"u}l et~al.(2025)Akg{\"u}l, Kalayc{\i}, Kannan, Neiswanger, and
  Prasanna}]{akgul2025lynx}
{\"O}mer~Faruk Akg{\"u}l, Yusuf~Hakan Kalayc{\i}, Rajgopal Kannan, Willie
  Neiswanger, and Viktor Prasanna. 2025.
\newblock \href {https://doi.org/10.48550/arXiv.2512.05325} {{LYNX}: Learning
  dynamic exits for confidence-controlled reasoning}.
\newblock arXiv preprint arXiv:2512.05325.
\newblock \emph{Preprint}, arXiv:2512.05325.

\bibitem[{Boppana et~al.(2026)Boppana, Ma, Loeffler, Sarfati, Bigelow, Geiger,
  Lewis, and Merullo}]{boppana2026reasoning}
Siddharth Boppana, Annabel Ma, Max Loeffler, Raphael Sarfati, Eric Bigelow,
  Atticus Geiger, Owen Lewis, and Jack Merullo. 2026.
\newblock \href {https://doi.org/10.48550/arXiv.2603.05488} {Reasoning theater:
  Disentangling model beliefs from chain-of-thought}.
\newblock arXiv preprint arXiv:2603.05488, version 4.

\bibitem[{Clark et~al.(2018)Clark, Cowhey, Etzioni, Khot, Sabharwal, Schoenick,
  and Tafjord}]{clark2018arc}
Peter Clark, Isaac Cowhey, Oren Etzioni, Tushar Khot, Ashish Sabharwal, Carissa
  Schoenick, and Oyvind Tafjord. 2018.
\newblock \href {https://doi.org/10.48550/arXiv.1803.05457} {Think you have
  solved question answering? try {ARC}, the {AI2} reasoning challenge}.
\newblock arXiv preprint arXiv:1803.05457.

\bibitem[{Dong et~al.(2026)Dong, Qin, and Shah}]{dong2026when}
Zhe Dong, Fang Qin, and Manish Shah. 2026.
\newblock \href {https://doi.org/10.48550/arXiv.2606.30852} {When does learning
  to stop help? a cost-aware study of early exits in reasoning models}.
\newblock arXiv preprint arXiv:2606.30852, version 2.

\bibitem[{Fang et~al.(2025)Fang, Ma, and Wang}]{fang2025thinkless}
Gongfan Fang, Xinyin Ma, and Xinchao Wang. 2025.
\newblock \href {https://doi.org/10.52202/085713-5060} {Thinkless: {LLM} learns
  when to think}.
\newblock In \emph{Advances in Neural Information Processing Systems},
  volume~38, pages 167994--168021.

\bibitem[{Fu et~al.(2026)Fu, Peng, Bai, Qin, Li, Liang, Chen, Li, and
  Ouyang}]{fu2026blade}
Keshu Fu, Keqin Peng, Jun Bai, Shuhan Qin, Chen Li, Junzhu Liang, Yefei Chen,
  Jiaqi Li, and Yuanxin Ouyang. 2026.
\newblock \href {https://doi.org/10.48550/arXiv.2607.28966} {{BLADE}:
  Boundary-expanded and layer-adaptive dynamic exit for efficient {LLM}
  reasoning}.
\newblock arXiv preprint arXiv:2607.28966.
\newblock \emph{Preprint}, arXiv:2607.28966.

\bibitem[{Fu et~al.(2025)Fu, Chen, Zhu, Fu, Dai, Zhuang, Ma, Qiao, Rosing,
  Stoica, and Zhang}]{fu2025certaindex}
Yichao Fu, Junda Chen, Siqi Zhu, Zheyu Fu, Zhongdongming Dai, Yonghao Zhuang,
  Yian Ma, Aurick Qiao, Tajana Rosing, Ion Stoica, and Hao Zhang. 2025.
\newblock \href {https://doi.org/10.52202/085713-4751} {Efficiently scaling
  {LLM} reasoning programs with {Certaindex}}.
\newblock In \emph{Advances in Neural Information Processing Systems},
  volume~38, pages 157566--157604.

\bibitem[{Hassid et~al.(2025)Hassid, Synnaeve, Adi, and
  Schwartz}]{hassid2025dontoverthink}
Michael Hassid, Gabriel Synnaeve, Yossi Adi, and Roy Schwartz. 2025.
\newblock \href {https://doi.org/10.48550/arXiv.2505.17813} {Don't overthink
  it. preferring shorter thinking chains for improved {LLM} reasoning}.
\newblock \emph{Preprint}, arXiv:2505.17813.
\newblock Version 2, revised February 3, 2026.

\bibitem[{Kwon et~al.(2023)Kwon, Li, Zhuang, Sheng, Zheng, Yu, Gonzalez, Zhang,
  and Stoica}]{kwon2023pagedattention}
Woosuk Kwon, Zhuohan Li, Siyuan Zhuang, Ying Sheng, Lianmin Zheng, Cody~Hao Yu,
  Joseph~E. Gonzalez, Hao Zhang, and Ion Stoica. 2023.
\newblock \href {https://doi.org/10.1145/3600006.3613165} {Efficient memory
  management for large language model serving with {PagedAttention}}.
\newblock In \emph{Proceedings of the 29th Symposium on Operating Systems
  Principles}, SOSP '23, pages 611--626. Association for Computing Machinery.

\bibitem[{Li et~al.(2026)Li, Hou, Zhu, Feng, Ye, Lei, Chen, Chen, and
  Du}]{li2026adaptive}
Pingzhi Li, Bairu Hou, Yun Zhu, Yihao Feng, Ke~Ye, Tao Lei, Zhifeng Chen,
  Tianlong Chen, and Xianzhi Du. 2026.
\newblock \href
  {https://proceedings.iclr.cc/paper_files/paper/2026/hash/96dd7b14cde7e0a5194eceedafb2f249-Abstract-Conference.html}
  {Adaptive thinking: Large language models know when to think in latent
  space}.
\newblock In \emph{The Fourteenth International Conference on Learning
  Representations}.

\bibitem[{Liu et~al.(2023)Liu, Liu, Cui, Teng, Duan, Zhou, and
  Zhang}]{liu2023logiqa}
Hanmeng Liu, Jian Liu, Leyang Cui, Zhiyang Teng, Nan Duan, Ming Zhou, and Yue
  Zhang. 2023.
\newblock \href {https://doi.org/10.1109/TASLP.2023.3293046} {{LogiQA} 2.0---an
  improved dataset for logical reasoning in natural language understanding}.
\newblock \emph{IEEE/ACM Transactions on Audio, Speech, and Language
  Processing}, 31:2947--2962.

\bibitem[{Liu et~al.(2026)Liu, Liu, Yang, Wang, Zhang, Feng, Zhang, and
  Wang}]{liu2026neat}
Kang Liu, YongKang Liu, Xiaocui Yang, Peidong Wang, Wen Zhang, Shi Feng, Yifei
  Zhang, and Daling Wang. 2026.
\newblock \href {https://doi.org/10.18653/v1/2026.findings-acl.1231} {{NEAT}:
  Neuron-based early exit for large reasoning models}.
\newblock In \emph{Findings of the Association for Computational Linguistics:
  ACL 2026}, pages 24616--24627. Association for Computational Linguistics.

\bibitem[{Liu and Wang(2025)}]{liu2025answer}
Xin Liu and Lu~Wang. 2025.
\newblock \href {https://doi.org/10.18653/v1/2025.emnlp-main.904} {Answer
  convergence as a signal for early stopping in reasoning}.
\newblock In \emph{Proceedings of the 2025 Conference on Empirical Methods in
  Natural Language Processing}, pages 17896--17907. Association for
  Computational Linguistics.

\bibitem[{Muennighoff et~al.(2025)Muennighoff, Yang, Shi, Li, Fei-Fei,
  Hajishirzi, Zettlemoyer, Liang, Cand{\`e}s, and
  Hashimoto}]{muennighoff2025s1}
Niklas Muennighoff, Zitong Yang, Weijia Shi, Xiang~Lisa Li, Li~Fei-Fei,
  Hannaneh Hajishirzi, Luke Zettlemoyer, Percy Liang, Emmanuel Cand{\`e}s, and
  Tatsunori Hashimoto. 2025.
\newblock \href {https://doi.org/10.18653/v1/2025.emnlp-main.1025} {{s1}:
  Simple test-time scaling}.
\newblock In \emph{Proceedings of the 2025 Conference on Empirical Methods in
  Natural Language Processing}, pages 20275--20321. Association for
  Computational Linguistics.

\bibitem[{Palmer(2026)}]{palmer2026kev}
Jared Palmer. 2026.
\newblock \href
  {https://github.com/jaredpalmer/kev/tree/0fe8fc97c2bcc247fa3efb6e5c32af4e99770e91}
  {{Kev}}.
\newblock Software repository.
\newblock Source commit 0fe8fc9; accessed 2026-10-11.

\bibitem[{{Qwen Team}(2026)}]{qwen2026model}
{Qwen Team}. 2026.
\newblock \href {https://huggingface.co/Qwen/Qwen3.5-4B} {{Qwen3.5-4B}}.
\newblock Model card.
\newblock Accessed 2026-10-11.

\bibitem[{Snell et~al.(2025)Snell, Lee, Xu, and Kumar}]{snell2025scaling}
Charlie Snell, Jaehoon Lee, Kelvin Xu, and Aviral Kumar. 2025.
\newblock \href
  {https://proceedings.iclr.cc/paper_files/paper/2025/hash/1b623663fd9b874366f3ce019fdfdd44-Abstract-Conference.html}
  {Scaling {LLM} test-time compute optimally can be more effective than scaling
  parameters for reasoning}.
\newblock In \emph{The Thirteenth International Conference on Learning
  Representations}.

\bibitem[{Tafjord et~al.(2021)Tafjord, Dalvi, and
  Clark}]{tafjord2021proofwriter}
Oyvind Tafjord, Bhavana Dalvi, and Peter Clark. 2021.
\newblock \href {https://doi.org/10.18653/v1/2021.findings-acl.317}
  {{ProofWriter}: Generating implications, proofs, and abductive statements
  over natural language}.
\newblock In \emph{Findings of the Association for Computational Linguistics:
  ACL-IJCNLP 2021}, pages 3621--3634. Association for Computational
  Linguistics.

\bibitem[{Tu et~al.(2025)Tu, Lin, Zhang, Tian, Li, Lan, and
  Zhao}]{tu2025autothink}
Songjun Tu, Jiahao Lin, Qichao Zhang, Xiangyu Tian, Linjing Li, Xiangyuan Lan,
  and Dongbin Zhao. 2025.
\newblock \href {https://doi.org/10.52202/085713-0510} {Learning when to think:
  Shaping adaptive reasoning in {R1}-style models via multi-stage {RL}}.
\newblock In \emph{Advances in Neural Information Processing Systems},
  volume~38, pages 16989--17015.

\bibitem[{Wang et~al.(2023)Wang, Wei, Schuurmans, Le, Chi, Narang, Chowdhery,
  and Zhou}]{wang2023selfconsistency}
Xuezhi Wang, Jason Wei, Dale Schuurmans, Quoc~V. Le, Ed~H. Chi, Sharan Narang,
  Aakanksha Chowdhery, and Denny Zhou. 2023.
\newblock \href {https://openreview.net/forum?id=1PL1NIMMrw} {Self-consistency
  improves chain of thought reasoning in language models}.
\newblock In \emph{The Eleventh International Conference on Learning
  Representations}.

\bibitem[{Wei et~al.(2022)Wei, Wang, Schuurmans, Bosma, Ichter, Xia, Chi, Le,
  and Zhou}]{wei2022chain}
Jason Wei, Xuezhi Wang, Dale Schuurmans, Maarten Bosma, Brian Ichter, Fei Xia,
  Ed~H. Chi, Quoc~V. Le, and Denny Zhou. 2022.
\newblock \href {https://doi.org/10.52202/068431-1800} {Chain-of-thought
  prompting elicits reasoning in large language models}.
\newblock In \emph{Advances in Neural Information Processing Systems},
  volume~35, pages 24824--24837.

\bibitem[{Xiang et~al.(2026)Xiang, Ji, Xu, Qiao, Yang, Li, and
  Zhang}]{xiang2026thinking}
Yang Xiang, Yixin Ji, Ruotao Xu, Dan Qiao, Zheming Yang, Juntao Li, and Min
  Zhang. 2026.
\newblock \href {https://doi.org/10.18653/v1/2026.acl-long.1080} {When is
  thinking enough? early exit via sufficiency assessment for efficient
  reasoning}.
\newblock In \emph{Proceedings of the 64th Annual Meeting of the Association
  for Computational Linguistics (Volume 1: Long Papers)}, pages 23541--23556.
  Association for Computational Linguistics.

\bibitem[{Yang et~al.(2025)Yang, Si, Duan, Zhu, Zhu, Li, Chen, Lin, and
  Wang}]{yang2025dynamic}
Chenxu Yang, Qingyi Si, Yongjie Duan, Zheliang Zhu, Chenyu Zhu, Qiaowei Li,
  Minghui Chen, Zheng Lin, and Weiping Wang. 2025.
\newblock \href {https://doi.org/10.48550/arXiv.2504.15895} {Dynamic early exit
  in reasoning models}.
\newblock arXiv preprint arXiv:2504.15895, version 3.

\bibitem[{Zhang et~al.(2025)Zhang, Lin, Hou, Feng, and
  Li}]{zhang2025adaptthink}
Jiajie Zhang, Nianyi Lin, Lei Hou, Ling Feng, and Juanzi Li. 2025.
\newblock \href {https://doi.org/10.18653/v1/2025.emnlp-main.184}
  {{AdaptThink}: Reasoning models can learn when to think}.
\newblock In \emph{Proceedings of the 2025 Conference on Empirical Methods in
  Natural Language Processing}, pages 3716--3730. Association for Computational
  Linguistics.

\end{thebibliography}

\appendix
\renewcommand{\topfraction}{0.9}
\renewcommand{\dbltopfraction}{0.9}
\renewcommand{\bottomfraction}{0.8}
\renewcommand{\textfraction}{0.08}
\renewcommand{\floatpagefraction}{0.8}
\renewcommand{\dblfloatpagefraction}{0.8}
\section{Reproducibility Specification}
\label{sec:reproducibility}
The supplementary materials at \url{https://github.com/binzheng-joey/selective-reasoning} contain code, reader weights, control parameters, evaluation manifests, and per-question measurements. A CPU script recomputes the reported statistics from these measurements; model inference requires the external model, source data, and feature caches described in the README. Table~\ref{tab:study-index} summarizes the experimental studies.

\begin{table*}[!tbp]
\centering\small
\setlength{\tabcolsep}{3pt}
\begin{tabular*}{\textwidth}{@{\extracolsep{\fill}}p{0.19\textwidth}p{0.18\textwidth}p{0.22\textwidth}p{0.10\textwidth}p{0.24\textwidth}@{}}
\toprule
Study & Questions/tasks & Answer interface & Token cap & Evidence role and backend\\
\midrule
Primary comparison & 768; PW/LQ/ARC & Current answer-position head; generated-answer adaptations & 0--8,192 & Frozen online policies; compiled A100 vLLM\\
Matched and supplementary policies & Same observed 768; fixed 48 for repeated time & Current head and deployed coefficients & 1,024 & Local component controls and follow-up; A100 vLLM\\
Reader/trigger development & 1,024 train; 256 observed development; PW/LQ & Matched native-state and answer-cue heads & 1,024 & Fitting and fixed-trajectory analysis; cached readouts\\
Earlier native confirmation & Separate 768; PW/LQ & Earlier linear head and native candidate logits & 1,024/8,192 & Frozen component comparisons; A100 vLLM\\
Kev routing & Separate 512; PW/LQ & Kev Base head; external native generator & 8,192 & Two-model component study; A100 vLLM\\
\bottomrule
\end{tabular*}
\caption{Study identities and evidence roles. PW, LQ, and ARC denote ProofWriter, LogiQA~2.0, and ARC-Challenge.}
\label{tab:study-index}
\end{table*}

\subsection{Data and Material Isolation}
ProofWriter uses the open-world, depth-5 files of release V2020.12.3; LogiQA~2.0 and ARC-Challenge use the source revisions recorded in the archive. ProofWriter and LogiQA evaluation questions come from official training data, and ARC from official test data. Reader fitting and observed development contain 512 and 128 questions per task from ProofWriter and LogiQA.

Material groups are connected components sharing source IDs within a dataset, normalized material, normalized sentence sets, or five-word-shingle Jaccard similarity of at least 0.90. Normalization lowercases and joins \texttt{\textbackslash w+} matches; sentence sets split at \texttt{[.!?]+} and deduplicate normalized sentences. Training excludes previously observed components, components spanning official splits, and conflicting source IDs. All prefixes inherit their question's component.

Primary evaluation selects one question per unused component after propagating exclusions from all recorded training, development, and earlier evaluations, including the earlier 1,024-question native confirmation pool. Deterministic hash ranking selects 64 components per ProofWriter subtype and 256 LogiQA components. The archive retains the ranking rules and selected IDs.

ARC groups questions by case-folded text with collapsed whitespace and selects 256 groups by deterministic hash ranking. Development excludes official training or validation questions whose normalized text appears in official test data. Source labels and candidate order are preserved. The evaluation contains 255 four-option items and one five-option item, \nolinkurl{TIMSS_1995_8_N4}.

The archive README locates data splits, material-group manifests, reader folds, deployed weights, and aggregate results. \nolinkurl{arr_formal_run.py} defines prompt projection, \nolinkurl{arr_integrated_online.py} implements the controller, and \nolinkurl{sentence_online.py} provides the reader backend. The evaluation manifest records all 768 question IDs and strata; separate protocols identify matched and supplementary executions.

\subsection{Generation and Answer-Query Templates}
All policies start from the same task-specific generation prompt. ProofWriter material is the original theory, its task is the selected statement, and its candidates are \texttt{True}, \texttt{False}, and \texttt{Unknown}, in that order. LogiQA material, question, and options retain their source fields. ARC material is the science question and its task is the exact string \texttt{Select the candidate that best answers this science question.}

The system message for ProofWriter and LogiQA is:
\begin{quote}\small\ttfamily
Use only the supplied material. Do not invent missing facts. Keep the original candidate order. Follow the requested output format.
\end{quote}
The ARC system message is:
\begin{quote}\small\ttfamily
Use the supplied question and relevant scientific knowledge. Keep the original candidate order. Follow the requested output format.
\end{quote}
The user-message template has one numbered line per candidate. Braced fields are substitutions; the displayed ellipsis expands to the intervening candidates, and blank lines are two newline characters.
\begin{quote}\small\ttfamily
Material:\\
\{material\}

Task:\\
\{task\}

Candidates (in the given order):\\
1. \{candidate 1\}\\
\ldots\\
\{K\}. \{candidate K\}

\{generation instruction\}
\end{quote}
For ProofWriter and LogiQA, the generation instruction is:
\begin{quote}\small\ttfamily
Reason through the task using only the material. Then give exactly one final candidate index from 1 to \{K\} in the exact format <FINAL>index</FINAL>. Place this final tag after your reasoning, outside any thinking block. Do not append text after the tag.
\end{quote}
For ARC, it is:
\begin{quote}\small\ttfamily
Reason through the question using the supplied information and relevant scientific knowledge. Then give exactly one final candidate index from 1 to \{K\} in the exact format <FINAL>index</FINAL>. Place this final tag after your reasoning, outside any thinking block. Do not append text after the tag.
\end{quote}
The checkpoint's chat template uses \nolinkurl{tokenize=True}, \nolinkurl{add_generation_prompt=True}, and \nolinkurl{enable_thinking=True}. Selected inputs are not truncated; the longest has 449 tokens, within the 20,000-token context with the largest output allowance.

An answer query concatenates original prompt and prefix token IDs with a suffix encoded using \nolinkurl{add_special_tokens=False}: \verb|\n\n<FINAL>| after a thinking closure, or \verb|\n</think>\n\n<FINAL>| otherwise. Here \verb|\n| denotes a newline. The prefix is not re-encoded. The reader uses the suffix's final state rather than its auxiliary output token.

Candidate $j$ maps to head row and stored prediction $j-1$. For $K\leq4$, inference casts the state to FP32 and applies the first $K$ head rows and softmax in FP32. All shared-reader policies use native logits over candidate-index tokens for the five-option ARC item, whose token IDs are 16--20.

Forced final completions use the same suffix and a 64-token cap, stopping at \texttt{</FINAL>} with the stop string included. Parsing prepends the injected \texttt{<FINAL>} cue and accepts a candidate index in final tags, \texttt{\textbackslash boxed\{...\}}, or a sole numeric response. Conflicting or out-of-range indices are invalid. Native parsing uses text after the last thinking closure and removes trailing chat EOS; a missing closure or valid answer is an error.

DEER trials append \verb|\n**Final Answer**\n| and \verb|The final candidate index is \boxed| to the original prompt and re-encoded history, excluding the triggering \texttt{Wait}. Rejection discards trial output, appends \texttt{Wait} to the saved history, and resumes reasoning. A trial ends at \texttt{</think>}; confidence is zero unless this closure is present and the output has at least two tokens. Otherwise confidence is the geometric mean of probabilities after the first token, including the closure. Acceptance requires confidence strictly above the threshold. A separate forced completion supplies the final answer after acceptance or terminal reasoning.

\subsection{Selected Reader and Control Parameters}
The seed-29 reader has 10,240 FP32 parameters, fitted with AdamW, weight decay 0.01, 16 questions per batch, and norm-1 gradient clipping. Each question has equal weight across capped positions $0,128,256,512,1024$. Development negative log-likelihood selects epoch 20 and learning rate $3\times10^{-4}$ from epochs 5, 10, and 20 and rates $3\times10^{-4}$ and $10^{-3}$. The backbone is frozen.

Four deterministic material-group folds supply out-of-fold training readouts. Each fold holds out 128 questions per task; its reader trains on the other 768 questions with seed 29, 20 epochs, and learning rate $3\times10^{-4}$. The full-data reader serves development and online evaluation.

The benefit gate has 2,564 inputs defined in Section~\ref{sec:method}. Per task, 384 training components fit the event models, 64 support probability-calibrated development variants, and 64 set routing thresholds. The two L2 logistic models use fitting-only standardization, $C=0.01$, LBFGS, 3,000 maximum iterations, tolerance $10^{-7}$, and seed 20261009. Standardization is folded into deployed coefficients.

The 0.75 routing threshold is the midpoint between descending held-out scores 96 and 97 among 128 questions. The raw-benefit threshold is 0.002275481505484662; the negative-margin threshold is $-0.8598166108131409$. Both use strict greater-than comparisons. The deployed benefit score uses uncalibrated logistic probabilities.

Trigger fitting samples at most two complete-sentence prefixes per question in each of four 256-token bands. An event receives weight $n_b/(m_b n)$, where $n_b$ and $m_b$ are available and sampled band counts and $n$ is the question's available-event count. Terminal-only events are excluded; common observable positions coalesce to the latest complete prefix, with weights renormalized within question. Development retains every available sentence event.

Training-only standardization and randomized PCA retain 64 natural-state dimensions with seed 20261009. The taskless trigger uses weighted L2 logistic regression at $C=0.01$ with the gate's solver and fitting limits. Its features are the projected state and $t/1024$; its out-of-fold target is a correct answer with certainty at least 0.85. Folding the transformations gives Eq.~\ref{eq:trigger}. Online evaluation clips the logit to $[-80,80]$, requests reads at $r_t\geq0.25$, and applies stopping certainty 0.70, 0.85, or 0.95 separately.

\subsection{Causal Sentence Events and State Availability}
The incremental detector reports each newly observable sentence boundary once. Chinese sentence punctuation ends a sentence immediately; English \texttt{.}, \texttt{!}, and \texttt{?}, optionally followed by closing quotes, require following whitespace. Filters exclude decimals, ellipses, enumerations, and abbreviations. Decoding preserves special tokens and spacing, deferring incomplete characters. Boundaries align to the earliest covering original token and exclude tokens containing subsequent non-whitespace content.

Each boundary increments the absolute sentence ordinal and replaces the pending event. The learned schedule tests $s\bmod4=0$; scored events are consumed once. New events can replace a pending prefix while a submitted read stays bound to its original prefix. State waiting, completion, and cancellation follow Table~\ref{alg:controller}.

\subsection{Software, Precision, and Execution Limits}
Online execution uses Python 3.11.13, vLLM 0.17.1, PyTorch 2.10.0/CUDA 12.8, Transformers 4.57.6, Triton 3.6.0, and NumPy 2.2.6 on A100 80GB PCIe GPUs. The backbone uses BF16, the reader and trigger FP32, and TF32 matrix multiplication is disabled. Fitting replay uses BF16 SDPA with FP32 readout logits.

\begin{table}[!htbp]
\small
\begin{tabular*}{\columnwidth}{@{\extracolsep{\fill}}p{0.56\columnwidth}p{0.37\columnwidth}@{}}
\toprule
Engine field & Value\\
\midrule
Tensor parallel size & 1\\
Maximum model length & 20,000\\
Maximum scheduled sequences & 2\\
Maximum batched tokens & 2,048\\
GPU memory utilization & 0.30\\
Attention backend & \nolinkurl{TRITON_ATTN}\\
Executor backend & \nolinkurl{uni}\\
Language-model-only mode & Enabled\\
Asynchronous scheduler & Disabled\\
Eager execution & Disabled\\
CUDA graph capture sizes & 1, 2\\
Cross-request prefix cache & Disabled\\
Generation configuration & \nolinkurl{vllm}\\
Engine and request seed & 20261009\\
\bottomrule
\end{tabular*}
\caption{Common online engine configuration. Generation and reading remain separate scheduled requests despite the disabled asynchronous engine scheduler.}
\label{tab:engine-specification}
\end{table}

Requests use temperature 0, top-$p$ 1, top-$k$ $-1$, zero presence and frequency penalties, and repetition penalty 1. Token IDs are returned without engine detokenization. Natural EOS can precede the policy cap. Reader requests emit one token; Answer Consistency trials and forced final completions have separate 64-token caps, and DEER permits at most ten 20-token trials. All emitted tokens count as output. The controller has a 300-second per-question limit.

One question--policy pair runs at a time, with at most one reasoning request and one answer query sharing the engine. Request-local KV state is retained, and between-policy cache resets precede timing. Appendix~\ref{sec:cost-details} defines the synchronized timing interval, repeated-run aggregation, and work counters.

\FloatBarrier
\section{Comparison Policy Details}
\label{sec:baseline-settings}
\paragraph{Answer interfaces and online adaptations.}
Shared-reader policies use the answer-position reader of Section~\ref{sec:method}. Forced terminal answering and Answer Consistency trials generate completions after \texttt{<FINAL>}. The native 8,192-token reference generates reasoning and its final answer without a forced cue. Invalid answers count as errors.

Answer Consistency (AC) adapts Answer Convergence to candidate answers, using $k=10$ consecutive matching trials at causal sentence boundaries. It pauses, queries, and resumes serially; repeated prefix processing and trial tokens are charged. Invalid or conflicting answers reset agreement. The DEER adaptation uses \texttt{Wait} transitions, geometric-mean trial confidence, a thinking-end requirement, and rollback after rejection. Greedy trials use threshold 0.95, at most ten 20-token trials, and a separate 64-token final-answer cap. This adaptation executes trials serially, without DEER's parallel branch variant.

\paragraph{Baseline execution behavior.}
\label{sec:baseline-behavior}
At caps of 1,024 and 2,048, primary DEER executions make 587 and 2,638 trial queries across 768 questions. The largest trial confidence is 0.8608, below the 0.95 acceptance threshold. Their 655 and 673 correct answers therefore come from terminal completion; development-selected thresholds are evaluated in Appendix~\ref{sec:calibrated-policies}.

Answer Consistency stops early on 765/768 questions and answers 522 correctly, including 105/256 on ProofWriter. Native generation answers 574/768 correctly; 147 traces lack a thinking closure and five lack a valid final choice. In total, 151 executions reach the output cap. All 152 format failures remain in the accuracy denominator.

\paragraph{Matched controls.}
The six conditions share a 1,024-token cap, reader, and $\tau=0.70$ and interleave within question on the same GPU. Always reason retains initial reading and gate evaluation before forcing generation. Margin gate replaces the benefit score. Fixed reads replaces the learned schedule with a gap of at least eight sentences since the last submitted read, initializing its ordinal to zero and retaining latest-prefix handling. No early stop retains intermediate queries but rejects early acceptance. Gate + terminal read retains routing and removes intermediate reads. Each comparison uses its matched complete-controller execution.

\paragraph{State and query interfaces.}
\label{sec:controller-interface}
The decoding hook indexes the 2,560-dimensional natural state $h_t$ by prefix length $t$ and removes states more than 64 tokens behind the latest captured position. A sentence boundary may precede its state's availability. An eligible event waits for that state until a newer event replaces it or generation ends. Capture, retention, and synchronization are included in pipeline time.

A query snapshots the original prompt and selected prefix token IDs, then appends the answer cue specified in Appendix~\ref{sec:reproducibility}. Its distribution and one auxiliary output token remain bound to that prefix while generation continues.

\begin{table}[!htbp]
\small
\begin{tabular*}{\columnwidth}{@{\extracolsep{\fill}}p{\columnwidth}@{}}
\toprule
\textbf{Algorithm: selective reasoning with concurrent reads}\\
\midrule
1. Read $(a_0,p_0,\tilde h_0)$; return $a_0$ if $u(x)\leq\theta$.\\
2. Start generation with cap $B$; initialize an empty pending prefix and no answer query.\\
3. On each sentence event, replace the pending prefix by $(z_{1:t},s)$. Generation remains scheduled.\\
4. If generation is active, the reader is idle, and a prefix is pending, discard it when $s\bmod4\ne0$. Otherwise retain it until its indexed $h_t$ is available or a newer event replaces it.\\
5. With generation active and the reader idle, consume an eligible pending event once $h_t$ is available. Evaluate $r_t$ once; if $r_t\geq0.25$, submit a read bound to that prefix, otherwise await a new event.\\
6. On read completion, accept if it is terminal or $q_t\geq\tau$. Cancel unfinished generation and return its $a_t$; otherwise continue with the latest pending event.\\
7. After EOS or budget exhaustion, finish any outstanding read. If no answer is accepted, request a terminal read and return through step~6.\\
\bottomrule
\end{tabular*}
\caption{The generator and reader are separate requests to one frozen backbone. Pending prefixes coalesce; at most one answer request runs at a time. A terminal read is accepted unconditionally.}
\label{alg:controller}
\end{table}

\section{Timing and Computational Work}
\label{sec:cost-details}
Primary costs average complete single-A100 executions on 768 questions under common model and engine settings. The timing check in Table~\ref{tab:same-device-timing} uses first executions on 48 of these questions for both accuracy and time. Table~\ref{tab:full-cohort-costs} supplements the mean costs with input processing, trigger calls, and latency tails.

% Generated by scripts/render_results.py; edit the renderer, not the cells.
\begin{table}[!tbp]
\centering\small
\setlength{\tabcolsep}{3.5pt}
\begin{tabular*}{\columnwidth}{@{}@{\extracolsep{\fill}}rrrrr@{}}
\toprule
$\tau$ & Acc. (\%) & Time (s) & Output tok. & Queries\\
\midrule
0.70 & 84.38 & 5.053 & 524.5 & 2.43\\
0.85 & 84.38 & 5.845 & 604.8 & 3.12\\
0.95 & 84.64 & 7.012 & 721.2 & 4.38\\
\bottomrule
\end{tabular*}
\caption{Acceptance-threshold operating points for the primary shared-reader controller on the same 768 questions and A100. The predefined primary threshold is 0.70; all three thresholds were fixed before evaluation. Accuracy is in percent; costs are per-question means including auxiliary queries.}
\label{tab:threshold-operating-points}
\end{table}

% Generated by scripts/render_results.py; edit the renderer, not the cells.
\begin{table*}[!tbp]
\centering\small
\setlength{\tabcolsep}{3.5pt}
\begin{tabular*}{\textwidth}{@{}@{\extracolsep{\fill}}lrrr@{}}
\toprule
Method & Acc. (\%) [95\% CI] & Mean s [95\% CI] & P95 s\\
\midrule
Direct read & 68.75 [56.25, 79.17] & 0.08 [0.08, 0.08] & 0.09\\
Terminal read (512) & 77.08 [66.67, 87.50] & 4.77 [4.76, 4.79] & 4.85\\
Terminal read (1,024) & 87.50 [79.17, 95.83] & 9.41 [9.40, 9.43] & 9.53\\
Terminal read (2,048) & 87.50 [79.17, 95.83] & 17.62 [16.98, 18.18] & 19.08\\
Predefined simple controller & 85.42 [75.00, 93.75] & 5.50 [4.47, 6.54] & 10.19\\
Ours ($\tau=0.70$) & 89.58 [81.25, 95.83] & 4.70 [3.84, 5.50] & 9.87\\
Ours ($\tau=0.85$) & 89.58 [81.25, 95.83] & 5.55 [4.62, 6.44] & 9.93\\
Ours ($\tau=0.95$) & 89.58 [81.25, 95.83] & 7.55 [6.42, 8.61] & 10.43\\
\midrule
Forced terminal answer & 87.50 [79.17, 95.83] & 9.46 [9.43, 9.49] & 9.66\\
Answer Consistency ($k=10$) & 66.67 [56.25, 77.08] & 4.22 [3.72, 4.72] & 8.45\\
DEER adaptation (1,024) & 83.33 [72.92, 91.67] & 9.68 [9.60, 9.76] & 10.18\\
DEER adaptation (2,048) & 87.50 [79.17, 95.83] & 18.98 [18.17, 19.75] & 22.64\\
Native answer (8,192) & 77.08 [66.67, 87.50] & 33.96 [28.41, 39.89] & 76.16\\
\bottomrule
\end{tabular*}
\caption{First-execution accuracy and timing on 48 preselected questions, 16 per task, with paired policy executions. Parenthesized numbers identify the policy caps. Pointwise 95\% bootstrap intervals quantify question sampling; three-run timing is analyzed separately.}
\label{tab:same-device-timing}
\end{table*}

% Generated by scripts/render_results.py; edit the renderer, not the cells.
\begin{table}[!tbp]
\centering\small
\setlength{\tabcolsep}{3.5pt}
\begin{tabularx}{\columnwidth}{@{}>{\raggedright\arraybackslash}Xrrr@{}}
\toprule
Method & P95 s & Input tok. & Triggers\\
\midrule
Direct read & 0.09 & 256 & 0.00\\
Terminal read (512) & 4.81 & 1,019 & 0.00\\
Terminal read (1,024) & 9.48 & 1,528 & 0.00\\
Terminal read (2,048) & 18.99 & 2,393 & 0.00\\
Predefined simple controller & 10.27 & 3,968 & 0.00\\
Ours ($\tau=0.70$) & 9.92 & 1,673 & 6.73\\
Ours ($\tau=0.85$) & 10.15 & 2,348 & 7.76\\
Ours ($\tau=0.95$) & 10.40 & 3,613 & 9.27\\
\midrule
Forced terminal answer & 9.82 & 1,528 & 0.00\\
Answer Consistency ($k=10$) & 9.05 & 10,506 & 0.00\\
DEER adaptation (1,024) & 10.21 & 3,093 & 0.00\\
DEER adaptation (2,048) & 22.72 & 13,546 & 0.00\\
Native answer (8,192) & 76.34 & 250 & 0.00\\
\bottomrule
\end{tabularx}
\caption{Tail latency and observation work on all 768 questions per method. P95 is the 95th percentile of pipeline seconds. Input tokens and trigger evaluations are per-question means; inputs include repeated prefix processing. Parenthesized numbers identify the policy caps.}
\label{tab:full-cohort-costs}
\end{table}

% Generated by scripts/render_results.py; edit the renderer, not the cells.
\begin{table*}[!tbp]
\centering\small
\setlength{\tabcolsep}{3.5pt}
\begin{tabular*}{\textwidth}{@{}@{\extracolsep{\fill}}lrrrrrrr@{}}
\toprule
Policy & Correct & Routed (\%) & Time (s) & Output tok. & Reads & Input tok. & Triggers\\
\midrule
Ours & 652 & 75.13 & 4.960 & 516.4 & 2.43 & 1,666.2 & 6.66\\
No early stop & 654 & 75.13 & 7.524 & 773.3 & 5.25 & 4,514.2 & 10.02\\
Gate + terminal read & 645 & 75.13 & 7.107 & 770.0 & 1.75 & 1,405.2 & 0.00\\
Always reason & 657 & 100.00 & 6.102 & 636.5 & 2.82 & 2,005.8 & 8.09\\
Margin gate & 655 & 64.71 & 4.838 & 503.6 & 2.43 & 1,686.9 & 6.52\\
Fixed reads & 643 & 75.13 & 4.554 & 462.6 & 4.11 & 2,894.4 & 0.00\\
\bottomrule
\end{tabular*}
\caption{Full matched component controls, including the margin-gate and fixed-reading replacements. Correct counts are out of 768; routed denotes the reasoning invocation rate. Time is mean pipeline seconds and Reads counts answer queries. Work metrics are per-question means. Inputs include repeated prefix processing; triggers count actual evaluations of the learned reading score.}
\label{tab:control-work}
\end{table*}

\paragraph{Timing scope.}
The synchronized timer spans the first read through the final decision and cancellation, including routing, generation, sentence detection, state capture, triggers, and answer or trial requests. Tokenization, loading, warm-up, cache resets, and record writing are excluded. Request-local KV state is retained and cross-request prefix caching is disabled, so input counts include repeated processing. Overlapping generation and read durations cannot be added to reconstruct latency.

\begin{table}[!htbp]
\small
\begin{tabular*}{\columnwidth}{@{\extracolsep{\fill}}p{0.24\columnwidth}p{0.68\columnwidth}@{}}
\toprule
Counter & Definition for one complete-controller execution\\
\midrule
Trigger calls & Actual evaluations of $r_t$ on eligible events with an available state. Sentence events and waiting steps do not increment this counter.\\
Answer requests & $N_{\mathrm{read}}=1+N_{\mathrm{mid}}+I_{\mathrm{fb}}$: initial read, all submitted intermediate reads, and an optional terminal fallback.\\
Input tokens & The generation prompt, if invoked, plus every answer-query input, including repeated prefixes and query suffixes.\\
Output tokens & $|z_{1:c}|+N_{\mathrm{read}}$, where $c$ is the actual natural output length and each read emits one auxiliary token.\\
\bottomrule
\end{tabular*}
\caption{Computational-work counters with cross-request prefix caching disabled. $I_{\mathrm{fb}}$ indicates a terminal fallback and $N_{\mathrm{mid}}$ counts intermediate reads. Token counts describe work separately from measured latency.}
\label{tab:controller-work}
\end{table}

An in-flight read accepted after EOS adds no fallback. Natural-generation requests are separate from answer requests; generation beyond the accepted prefix and after the stop decision is included in actual output.

\paragraph{Uncertainty and repeated executions.}
Online comparisons use pointwise 95\% paired material-group bootstrap intervals with 10,000 resamples, stratified by ProofWriter subtype, LogiQA, and ARC. Repeated timing averages three executions within each of 48 fixed questions before resampling. Repetitions reuse resident engines and contribute no additional independent accuracy subjects. Controls and timing checks retain within-series same-GPU pairing.

\begin{table*}[!tbp]
\centering\small
\setlength{\tabcolsep}{5pt}
\begin{tabular*}{\textwidth}{@{}@{\extracolsep{\fill}}llrr@{}}
\toprule
Comparison & Metric & Estimate & 95\% confidence interval\\
\midrule
\multicolumn{4}{l}{\textit{Primary executions: predefined controller versus shared-reader baselines}}\\
Ours versus direct read & Accuracy difference (pp) & $+13.41$ & $+10.42$ to $+16.28$\\
Ours versus terminal-512 & Accuracy difference (pp) & $+7.29$ & $+4.69$ to $+9.90$\\
Ours versus terminal-1,024 & Accuracy difference (pp) & $-1.17$ & $-2.60$ to $+0.26$\\
Ours versus terminal-1,024 & Time saved (\%) & 46.30 & 43.84 to 48.79\\
\midrule
\multicolumn{4}{l}{\textit{Matched executions: local complete-controller and component controls}}\\
Complete versus no early stop & Time saved (\%) & 34.09 & 31.88 to 36.28\\
Benefit versus margin gate & Accuracy difference (pp) & $-0.39$ & $-1.56$ to $+0.78$\\
Benefit versus margin gate & Time difference (s) & $+0.121$ & $-0.096$ to $+0.332$\\
Learned versus fixed reads & Accuracy difference (pp) & $+1.17$ & $-0.39$ to $+2.73$\\
Learned versus fixed reads & Time difference (s) & $+0.406$ & $+0.275$ to $+0.539$\\
\bottomrule
\end{tabular*}
\caption{Paired comparisons from the first executions on all 768 questions. The primary controller answers 648 questions correctly; the independently executed matched complete controller answers 652. Each comparison pairs executions on the same A100 within its series. Accuracy and time differences subtract the second policy from the first; pp denotes percentage points. Positive time savings indicate the reduction relative to the second policy. Intervals use the paired material-group bootstrap protocol in Appendix~\ref{sec:cost-details}.}
\label{tab:paired-comparisons}
\end{table*}

\section{Mechanism and Stability Details}
\label{sec:mechanism-details}
\paragraph{Routing by task.}
The matched direct and always-reason endpoints answer 541 and 657 questions correctly: 147 corrections, 31 regressions, 510 correct under both, and 80 wrong under both. Replaying each gate's choices over these endpoints reproduces its 768 deployed answers in Table~\ref{tab:gate-tasks}. These diagnostic endpoints differ from the terminal-1,024 endpoint used for gate training.

\begin{table}[!htbp]
\centering\small
\begin{tabular*}{\columnwidth}{@{\extracolsep{\fill}}lrrrr@{}}
\toprule
 & \multicolumn{2}{c}{Benefit gate} & \multicolumn{2}{c}{Margin gate}\\
Task & Calls & Correct & Calls & Correct\\
\midrule
ProofWriter & 240 & 211 & 235 & 210\\
LogiQA~2.0 & 162 & 199 & 179 & 202\\
ARC-Challenge & 175 & 242 & 83 & 243\\
\midrule
All & 577 & 652 & 497 & 655\\
\bottomrule
\end{tabular*}
\caption{Matched gate replacement on 256 questions per task, with the same GPU, downstream schedule, and stopping rule. Calls count reasoning invocations; correct counts include direct and reasoning-assisted answers.}
\label{tab:gate-tasks}
\end{table}

Benefit alone selects 160 questions: two corrections, one regression, and 157 unchanged outcomes. Margin alone selects 80: ten corrections, six regressions, and 64 unchanged outcomes. On ARC, 107 of the 108 questions uniquely selected by benefit have unchanged correctness, locating its excess allocation.

\paragraph{Reading opportunities.}
Post-hoc inspection matches 1,077 of 4,229 rejected trigger positions to exact-prefix reads from the fixed schedule. Among these matches, 121 questions have a correct rejected opportunity with certainty at least 0.70; 70 fall below the 0.85 training target. These observations motivate target alignment and bounded waiting.

\paragraph{Early-exit errors.}
Among 369 cancelled questions, 25 exits are wrong (6.78\%), but continued reasoning corrects only two (0.54\% of cancellations; 0.26\% of all 768 questions). Those two ARC executions stop at 104 and 237 tokens, compared with 1,024 and 972 without early stopping. The other 23 erroneous exits remain wrong after continuation.

\paragraph{Observation and cancellation work.}
The matched controller's initial query averages 0.0828 seconds; subsequent read-response durations sum to 0.1095 seconds per question while overlapping generation. Mean input processing is 1,666.18 tokens. Generation beyond an accepted prefix averages 0.961 tokens over all 768 questions; recorded tokens after the stop decision average zero.

\paragraph{Repeated and independent executions.}
All 13 primary configurations and six controls preserve answers and routing across their three executions on fixed 48-question subsets; one DEER-2,048 trace changes. Repeated timing gives a 45.48\% saving from early stopping, with a 95\% interval of 38.59--52.52\%, compared with the full-cohort estimate of 34.09\%.

The primary and matched controller executions answer 648 and 652 questions correctly. Their 35 final-answer, 11 routing, and 457 generation-sequence differences reflect independent runs. The margin/every-eight policy likewise answers 638 questions correctly in the factorial series and 648 in the calibrated series, despite the same policy and fitted components. Benefit-minus-margin correct counts differ by $-3$ in the matched comparison and $+2$ in the factorial comparison. Each reported contrast therefore uses its own local paired executions.

\section{Follow-up on Reading Opportunities}
\label{sec:trigger-revisions}
% Generated by scripts/render_results.py; edit the renderer, not the cells.
\begin{table*}[!tbp]
\centering\small
\setlength{\tabcolsep}{3.5pt}
\begin{tabular*}{\textwidth}{@{}@{\extracolsep{\fill}}lrrrr@{}}
\toprule
Schedule & Correct/768 & Output tok. & Queries & Time saved (\%) [95\% CI]\\
\midrule
Original trigger & 652 & 516.4 & 2.43 & Reference\\
Fixed every 8 sentences & 643 & 462.6 & 4.11 & 6.10 [-1.76, 13.76]\\
Aligned target & 642 & 456.4 & 3.52 & 13.56 [7.20, 20.56]\\
Read deadline & 646 & 464.6 & 3.62 & 11.62 [5.90, 17.38]\\
Aligned target + deadline & 643 & 437.5 & 4.27 & 15.07 [7.80, 22.78]\\
\bottomrule
\end{tabular*}
\caption{Post-diagnostic schedule revisions on the observed cohort. Correct counts and work use 768 questions; the two original schedules reuse the matched-control runs. Time savings use a fresh, interleaved five-policy comparison on 48 questions, averaging three executions per question before paired bootstrap. All schedules retain the start gate, 1,024-token cap, and stopping threshold 0.70. These revisions are separate from the primary policies in Table~\ref{tab:main-results}.}
\label{tab:trigger-revisions}
\end{table*}

We evaluate target alignment, a reading deadline, and their combination. Alignment refits the logistic trigger on the original training prefixes using $\mathbb{I}[a_t=y\land q_t\geq0.70]$, with the original projection and $C=0.01$. The deadline requests the latest complete prefix after the first eight sentences, or a gap of 16 sentences or 256 tokens since the last read. The reader, gate, 1,024-token cap, trigger threshold 0.25, and stopping threshold 0.70 remain fixed.

Table~\ref{tab:trigger-revisions} evaluates accuracy on the already observed 768 questions. The original and fixed schedules reuse their matched-control runs; the three revisions are new executions. All five schedules are rerun in interleaved order on the same GPU for repeated timing on 48 fixed questions, three times per question.

The original trigger leaves 104/577 routed questions without an intermediate read; alignment reduces this count to 30 and both deadline variants to zero. Among questions receiving a read, its first position averages 397 tokens for the original, 173 for fixed reads, 284 for alignment, 163 for the deadline, and 133 for their combination. These conditional populations differ; the revisions route 579 questions, compared with 577 in the original controls.

The combined revision and fixed reads each answer 643 questions correctly, with 17 corrections and 17 regressions between predictions and a 95\% accuracy-difference interval of $[-1.43,+1.43]$ points. The revision uses 437.5 rather than 462.6 output tokens and 4.27 rather than 4.11 reads; repeated time falls by 9.56\% (95\% interval: 2.61--16.80\%). Against the original trigger, it saves 15.07\% of repeated-subset time while full-cohort accuracy changes by $-1.17$ points (95\% interval: $[-2.73,+0.26]$). Earlier acquisition changes both cost and the answers returned.

\section{Supplementary Policy Comparisons}
\label{sec:supplementary-comparisons}
These supplementary comparisons retain the fitted backbone, reader, and controller weights and reuse the observed three-task cohort.

\subsection{Development-Selected Operating Points}
\label{sec:calibrated-policies}
% Generated by scripts/render_supplement.py; edit the renderer, not the cells.
\begin{table*}[!tbp]
\centering\small
\setlength{\tabcolsep}{3.5pt}
\begin{tabular*}{\textwidth}{@{}@{\extracolsep{\fill}}llrrrr@{}}
\toprule
\multicolumn{6}{l}{\textbf{(a) Accuracy and work: 768 questions}}\\
\midrule
Method & Stop criterion & Correct/768 & $\Delta$ acc. (pp) [95\% CI] & Output tok. & Reads\\
\midrule
Benefit gate + learned & $q_t\geq0.70$ & 648 & Reference & 523.0 & 2.44\\
Benefit gate + learned & $q_t\geq0.85$ & 648 & 0.00 [-0.39, 0.39] & 597.5 & 3.09\\
Benefit gate + learned & $q_t\geq0.95$ & 650 & 0.26 [-0.26, 0.78] & 709.6 & 4.33\\
Margin gate + every 8 & $q_t\geq0.70$ & 648 & 0.00 [-1.69, 1.69] & 481.3 & 4.26\\
Margin gate + every 8 & $q_t\geq0.85$ & 652 & 0.52 [-1.17, 2.34] & 587.0 & 5.09\\
Margin gate + every 16 & $q_t\geq0.95$ & 648 & 0.00 [-1.69, 1.82] & 656.1 & 3.49\\
DEER & $\tau_{\mathrm{trial}}=0.80$ & 649 & 0.13 [-1.69, 2.08] & 1021.9 & 1.74\\
DEER & $\tau_{\mathrm{trial}}=0.95$ & 649 & 0.13 [-1.69, 2.08] & 1026.4 & 1.74\\
Forced terminal answer & Terminal & 645 & -0.39 [-2.34, 1.56] & 1026.2 & 1.00\\
\bottomrule
\end{tabular*}
\par\medskip
\begin{tabular*}{\textwidth}{@{}@{\extracolsep{\fill}}llrr@{}}
\toprule
\multicolumn{4}{l}{\textbf{(b) Repeated timing: 48 questions}}\\
\midrule
Method & Stop criterion & Mean s & $\Delta$ mean s [95\% CI]\\
\midrule
Benefit gate + learned & $q_t\geq0.70$ & 4.123 & Reference\\
Benefit gate + learned & $q_t\geq0.85$ & 5.188 & 1.065 [0.542, 1.653]\\
Benefit gate + learned & $q_t\geq0.95$ & 7.123 & 3.000 [2.253, 3.724]\\
Margin gate + every 8 & $q_t\geq0.70$ & 3.981 & -0.142 [-0.997, 0.800]\\
Margin gate + every 8 & $q_t\geq0.85$ & 5.411 & 1.288 [0.193, 2.437]\\
Margin gate + every 16 & $q_t\geq0.95$ & 6.227 & 2.104 [0.921, 3.303]\\
DEER & $\tau_{\mathrm{trial}}=0.80$ & 9.554 & 5.431 [4.702, 6.200]\\
DEER & $\tau_{\mathrm{trial}}=0.95$ & 9.553 & 5.430 [4.699, 6.200]\\
Forced terminal answer & Terminal & 9.396 & 5.273 [4.570, 6.024]\\
\bottomrule
\end{tabular*}
\caption{Stopping-threshold comparisons reusing the 768 questions in Table~\ref{tab:main-results}. Benefit and margin denote the start gates; learned reads use the natural-state trigger, and every 8 or 16 denotes the sentence interval. All reasoning caps are 1,024 tokens. $q_t=1-H(p_t)/\log K$ measures answer-distribution concentration; DEER's $\tau_{\mathrm{trial}}$ is its separate trial threshold. Panel (a) uses the first execution on all 768 questions for accuracy and per-question work, including auxiliary outputs. Panel (b) uses 48 preselected questions, 16 per task, on one A100, averaging three executions within each question for time. The two panels have different denominators. Differences are method minus benefit gate with learned reads at $q_t\geq0.70$ within the respective cohorts. Intervals are pointwise 95\% paired stratified material-group bootstrap intervals.}
\label{tab:supplement-calibrated}
\end{table*}

The primary simple controller at cap 1,024 and $\tau=0.85$ reaches 81.64\% on ProofWriter, 79.30\% on LogiQA~2.0, 94.53\% on ARC-Challenge, and 85.16\% overall, with mean time 5.834 seconds, 591.2 output tokens, and 5.10 queries.

Operating-point selection uses 48 development questions, 16 per task and four per ProofWriter subtype. ProofWriter and LogiQA questions come from reader development; ARC questions come from official validation after excluding normalized groups appearing in official test data. All development IDs and recorded material groups are disjoint from evaluation.

At acceptance thresholds $0.70$, $0.85$, and $0.95$, the simple margin controller selects reading gaps from four, eight, or sixteen sentences by development correct count, breaking ties by latency. Selected gaps are eight, eight, and sixteen. DEER tests confidence thresholds $0.80$, $0.85$, $0.90$, and $0.95$; both the fastest and highest-accuracy development points with an early exit select $0.80$. Retaining the learned thresholds, default DEER, and forced terminal answering gives nine evaluation policies.

Each question's nine policies run on one GPU under common model and engine settings. Repeated timing averages three same-GPU executions within each of 48 preselected questions.

At $\tau=0.70$, learned and simple controllers both answer 648/768 correctly. Learned control reduces reads by 42.75\%, from 4.264 to 2.441, and input processing from 3,087 to 1,685 tokens, while output rises by 8.66\%, from 481.3 to 523.0. Their 95\% accuracy-difference interval is $[-1.69,+1.69]$ points. Simple-minus-learned repeated time differs by $-0.142$ seconds (95\% interval: $[-0.997,+0.800]$), leaving no resolved latency advantage.

DEER at $0.80$ exits early on eleven questions, compared with zero at $0.95$; both answer 649/768 correctly in this series. Its repeated time exceeds learned $0.70$ by 5.431 seconds (95\% interval: 4.702--6.200). Across 432 policy--question repetition groups, one changes its final answer and 54 change generated sequences.

\subsection{Gate and Reading Factorial}
\label{sec:factorial-controls}
% Generated by scripts/render_supplement.py; edit the renderer, not the cells.
\begin{table*}[!tbp]
\centering\small
\setlength{\tabcolsep}{3.5pt}
\begin{tabular*}{\textwidth}{@{}@{\extracolsep{\fill}}lllrrrr@{}}
\toprule
\multicolumn{7}{l}{\textbf{(a) Gate and read-schedule combinations: 768 questions}}\\
\midrule
ID & Start gate & Read schedule & Correct/768 & Mean s & Output tok. & Reads\\
\midrule
A & Benefit & Learned & 653 & 5.037 & 525.4 & 2.50\\
B & Benefit & Every 8 sentences & 640 & 4.631 & 471.6 & 4.18\\
C & Margin & Learned & 651 & 4.949 & 516.4 & 2.50\\
D & Margin & Every 8 sentences & 638 & 4.779 & 487.1 & 4.30\\
\bottomrule
\end{tabular*}
\par\medskip
\begin{tabular*}{\textwidth}{@{}@{\extracolsep{\fill}}lrrrr@{}}
\toprule
\multicolumn{5}{l}{\textbf{(b) Paired differences with 95\% confidence intervals}}\\
\midrule
Contrast & $\Delta$ acc. (pp) & $\Delta$ reads & \shortstack{$\Delta$ mean s\\768 questions} & \shortstack{$\Delta$ mean s\\48 questions $\times$ 3}\\
\midrule
$A-B$ & 1.69 [0.13, 3.39] & -1.681 [-1.823, -1.539] & 0.406 [0.277, 0.539] & 0.295 [-0.080, 0.689]\\
$A-C$ & 0.26 [-0.91, 1.43] & 0.000 [-0.083, 0.078] & 0.088 [-0.128, 0.300] & 0.490 [-0.244, 1.151]\\
$A-D$ & 1.95 [0.13, 3.91] & -1.805 [-1.977, -1.637] & 0.258 [0.019, 0.488] & 0.689 [-0.096, 1.416]\\
$B-C$ & -1.43 [-3.39, 0.52] & 1.681 [1.517, 1.842] & -0.318 [-0.534, -0.114] & 0.195 [-0.592, 0.916]\\
$B-D$ & 0.26 [-0.78, 1.30] & -0.124 [-0.250, -0.005] & -0.148 [-0.335, 0.026] & 0.393 [-0.276, 0.994]\\
$C-D$ & 1.69 [0.00, 3.39] & -1.805 [-1.949, -1.661] & 0.169 [0.051, 0.287] & 0.199 [-0.157, 0.584]\\
\bottomrule
\end{tabular*}
\caption{Factorial comparison of start gates and read schedules at $q_t\geq0.70$ and a 1,024-token reasoning cap. The four policies run on the same 768 questions from Table~\ref{tab:main-results} and physical A100 with per-question interleaving. Panel (a) reports first-execution correct counts and per-question means; output tokens include auxiliary requests. Panel (b) pairs the 768 first executions for accuracy, reads, and time, and also reports timing on 48 preselected questions, 16 per task, averaging three executions within each question before pairing. Each contrast subtracts the second policy from the first. Intervals are pointwise 95\% paired stratified material-group bootstrap intervals.}
\label{tab:supplement-factorial}
\end{table*}

Four combinations cross benefit versus margin routing with learned versus every-eight-sentence reading, using the same reader and acceptance $0.70$. All four interleave within question on the same GPU. Benefit obtains two additional correct answers under either reading schedule: $+0.26$ points, with 95\% intervals $[-0.91,+1.43]$ under learned reading and $[-0.78,+1.30]$ under fixed reading. Neither time interval resolves a difference.

With benefit routing fixed, learned reading obtains thirteen more correct answers and 40.24\% fewer reads, but time rises by 0.406 seconds (95\% interval: 0.277--0.539) and generation by 55.54 tokens. Margin routing shows the same direction of observation and generation changes. These factorial executions remain separate from primary and calibrated results.

\subsection{Serial and Concurrent Execution}
\label{sec:execution-comparison}
% Generated by scripts/render_supplement.py; edit the renderer, not the cells.
\begin{table}[!tbp]
\centering\small
\setlength{\tabcolsep}{3.5pt}
\begin{tabular*}{\columnwidth}{@{}@{\extracolsep{\fill}}lrrr@{}}
\toprule
\multicolumn{4}{l}{\textbf{(a) First execution: 48 questions}}\\
\midrule
Execution & Acc. (\%) [95\% CI] & Mean s & Input tok.\\
\midrule
Concurrent & 85.42 [75.00, 93.75] & 4.792 & 1664.2\\
Serial & 83.33 [72.92, 91.67] & 4.691 & 2022.7\\
\bottomrule
\end{tabular*}
\par\medskip
\begin{tabular*}{\columnwidth}{@{}@{\extracolsep{\fill}}lr@{}}
\toprule
\multicolumn{2}{l}{\textbf{(b) Serial minus concurrent}}\\
\midrule
Metric & Difference [95\% CI]\\
\midrule
Accuracy (pp) & -2.08 [-6.25, 0.00]\\
Time (s), first run & -0.101 [-0.276, 0.056]\\
Time (s), three-run mean & -0.103 [-0.277, 0.055]\\
Input tokens & 358.5 [16.4, 695.5]\\
Reads & -0.062 [-0.312, 0.104]\\
\bottomrule
\end{tabular*}
\caption{Serial and concurrent execution of the same controller on 48 preselected questions, 16 per task, with cap 1,024 and $\tau=0.70$. Panel (a) reports first-execution accuracy and mean costs. Panel (b) reports serial-minus-concurrent differences; the three-run time averages executions within question. Input tokens include serial prefix restoration. Intervals are pointwise 95\% stratified material-group bootstrap intervals, paired for differences.}
\label{tab:supplement-execution}
\end{table}

The controller executes on 48 questions with three same-GPU runs per mode. Serial reading aborts generation, reads the selected prefix, and resumes from the exact token prefix after rejection, charging each restored prefill. Concurrent reading keeps generation active while processing the snapshot.

Serial and concurrent execution answer 40/48 and 41/48 correctly on the first run. Their repeated serial-minus-concurrent time difference is $-0.103$ seconds (95\% interval: $[-0.277,+0.055]$). Serial input work rises by 21.54\%, with thirty restores across fifteen questions. Final sequences differ on 35/48 questions and the answer on one, so the comparison includes prefix-recovery and trajectory effects.

\section{Natural-State Information beyond Position}
\label{sec:position-diagnostic}
% Generated by scripts/render_position_diagnostic.py from completed CPU aggregates.
\begin{table*}[!tbp]
\centering\small
\begin{tabular*}{\textwidth}{@{}@{\extracolsep{\fill}}llrrrr@{}}
\toprule
Evaluation certainty & Score & Log loss & Pooled AUROC & AP & Within-question AUROC\\
\midrule
0.85 & Natural state + position & 0.371 & 0.805 & 0.527 & 0.746\\
0.85 & Position logistic & 0.466 & 0.671 & 0.289 & 0.793\\
0.85 & Position lookup & 0.448 & 0.665 & 0.269 & 0.782\\
\midrule
0.70 & Natural state + position & 0.582 & 0.758 & 0.640 & 0.756\\
0.70 & Position logistic & 0.663 & 0.670 & 0.457 & 0.843\\
0.70 & Position lookup & 0.626 & 0.666 & 0.439 & 0.832\\
\bottomrule
\end{tabular*}
\caption{Position-control diagnostic on 3,407 absolute modulo-four sentence checkpoints from 256 observed development questions, with 128 per task. The target is a correct answer read with entropy certainty at least the evaluation threshold. Every score is trained with the original 0.85 target; 0.70 is an additional evaluation without refitting. Pooled metrics give each question equal total weight. Within-question AUROC averages trajectories containing both target classes: 110/256 at 0.85 and 130/256 at 0.70. Position logistic is independently refitted on the original training questions; lookup uses fixed 64-token bins and training-only weighted rates. These cached events do not measure online latency.}
\label{tab:position-diagnostic}
\end{table*}

Saved trajectories from 1,024 training and 256 development questions in ProofWriter and LogiQA~2.0 compare the frozen taskless trigger with a position-only logistic model and a 64-token lookup. Both controls fit training data using the original out-of-fold target $\mathbb{I}[a_t=y\land q_t\geq0.85]$ and question weights. Logistic regression uses $t/1024$, $C=0.01$, and the trigger's fitting limits; lookup smooths weighted bin mass as $(m_++1)/(m+2)$.

The cache contains 8,189 sampled training events and 14,012 complete development events. Training sample indices are not sentence ordinals. Development ordinals come from the complete event list before coalescing observable positions; excluding terminal-only events and retaining $s\bmod4=0$ gives 3,407 candidates across 256 questions. Pooled weights normalize within question; 95\% intervals use 1,000 paired material-group bootstrap draws within task. Within-question analysis was added after inspecting pooled results.

At target 0.85, pooled AUROC is 0.805 for natural states, 0.671 for position logistic, and 0.665 for lookup. State-minus-lookup AUROC differs by 0.141 (95\% interval: 0.100--0.178) and log loss by $-0.077$ ($[-0.096,-0.059]$). Gains occur within both tasks. Conditioning all sentence events on 64-token bands gives AUROC 0.788 for states and 0.500 for lookup, supporting predictive information beyond position on this cache.

Within-question AUROC favors the position controls. Among 110 questions with both target classes, mean AUROC is 0.746 for states, 0.793 for position logistic, and 0.782 for lookup. State-minus-logistic differs by $-0.047$ (95\% interval: $[-0.0969,-0.0001]$) and state-minus-lookup by $-0.037$ ($[-0.088,+0.014]$). Evaluation at target 0.70 gives the same ordering. Pooled prediction and within-question temporal ranking measure different properties of the reading signal.

\paragraph{Fixed-trajectory reading outcomes.}
At stopping certainty 0.85 and trigger threshold 0.25, states and lookup both answer 203/256 correctly. State screening uses 2.543 rather than 3.465 reads, a difference of $-0.922$ (95\% interval: $[-1.195,-0.633]$), and 851.94 rather than 947.66 ideal tokens, a difference of $-95.72$ ($[-121.56,-69.30]$). Position logistic requests no intermediate reads at this threshold and falls back to terminal reading. Equal thresholds yield different query budgets.

Replay assumes instantaneous decisions on fixed prefixes, without an initial gate. Sentence distributions come from RTX3080 readouts and terminal distributions from the earlier A100 replay. These ideal-token outcomes omit online scheduling, state access, and cancellation costs; they are distinct from the three-task pipeline-time measurements.

\section{Training Details}
\label{sec:hyperparameters}
\paragraph{Readers.}
The native Qwen3.5-4B checkpoint uses \href{https://modelscope.cn/models/Qwen/Qwen3.5-4B}{ModelScope} revision \texttt{f09dfc75214c5f2f4940}\allowbreak\texttt{ae6a26c126ce93828a9b}. Its weight files match Hugging Face revision \texttt{851bf6e806efd8d0a36b}\allowbreak\texttt{00ddf55e13ccb7b8cd0a}. Bias-free linear heads map 2,560 dimensions to four candidates; pooling adds a learned 2,560-dimensional query. Seeds are 17, 29, and 43; learning rates are $\{3\times10^{-4},10^{-3}\}$. AdamW uses weight decay 0.01, batches of 16 questions, and norm-1 gradient clipping. Training lasts 20 epochs, with selection from epochs 5, 10, and 20 by development mean negative log-likelihood. Variation is standard deviation over selected seed runs.

\paragraph{Component gates.}
The Kev gate uses standardized 2,560-dimensional decision-token states and L2 logistic regression with LBFGS, 2,000 maximum iterations, and tolerance $10^{-7}$. Mean calibration Brier score across correction and regression heads selects $C=0.001$ from $\{0.001,0.01,0.1,1\}$.

The native-reader gate study fixes reader seed 29 and concatenates its initial answer state with maximum probability, margin, normalized entropy, and $\log(1+\text{input length})$. Gain, harm, and initial-error models use $C=0.01$ and 3,000 maximum iterations. Standardization uses fitting groups; global and task-conditional probability calibration and routing thresholds use separate groups.

\paragraph{Fixed-grid stopping pilot.}
Four material-group folds produce training predictions from seed-29 readers fitted on the other 768 questions for 20 epochs at $3\times10^{-4}$. Stop models use standardized features and L2 logistic regression with $C=1$. Features comprise candidate probabilities, maximum probability, margin, normalized entropy, budget fraction, candidate count, and performed-read history: previous confidence, confidence change, answer agreement, and whether a previous read exists.

Thresholds are $\{0.5,0.6,0.7,0.8,0.85,0.9,0.95,0.98,0.99\}$ and continuation prices are $\{0,0.025,0.05,0.1,0.2\}$. The joint target compares current and terminal correctness.

\paragraph{Sentence-trigger development.}
The development trigger uses out-of-fold answer distributions at sampled sentence prefixes, with equal total weight per question. Training-fitted standardization and 64-component PCA transform natural final-layer states. Logistic regression with $C=0.01$ receives this projection, generated-token fraction, and a binary task indicator. Standardization and projection fold into the affine trigger. The schedule study fixes entropy certainty at 0.85 and trigger probability at 0.25.

The three-task controller refits the trigger without the task indicator. Its training target remains correctness with entropy certainty at least 0.85, while the subsequent acceptance threshold varies independently.

\section{Component Study Protocols}
\label{sec:component-protocols}
\paragraph{Tasks and data separation.}
Component studies use separate ProofWriter and LogiQA~2.0 cohorts balanced between tasks \citep{tafjord2021proofwriter,liu2023logiqa}. Splits separate material groups. The Kev, native-component, and reader-development studies differ in endpoints, checkpoints, and backends; the three-task primary evaluation uses a distinct reserved cohort.

\paragraph{Kev routing study.}
The benefit gate uses 1,024 training and 256 calibration questions, then 512 questions from previously unused official-training material groups excluding historical project material and the reserved native-component cohort. Kev applies its trained adapter and decision head to Qwen3.5-4B-Base; an external post-trained Qwen3.5-4B generates reasoning before a second Kev decision. Generation uses BF16 vLLM and decisions use FP32. The cap is 8,192 tokens, reached by 106/512 traces.

\paragraph{Native component confirmation.}
A separate evaluation measures frozen native-backbone readers, routing rules, and fixed-grid stopping on 768 reserved material groups, 384 per task. The first 24 complete 32-question blocks were selected from the planned 1,024 before labels were accessed, retaining all 35 configurations and paired comparisons. The remaining 256 labels were unopened. These observations subsequently informed component development.

Native configurations use one A100 with BF16 vLLM and prefix caching disabled. Native reading normalizes language-model logits over valid answer-label tokens at the cue; linear reading applies a checkpoint distinct from the development reader, computing FP32 scores rounded to BF16 before normalization. Each question--policy pair runs separate generation and reading requests in seeded shuffled order. Checks use positions 0, 128, 256, 512, and 1,024. The long baseline generates to natural completion or 8,192 tokens, then performs a native answer read.

\paragraph{Matched readers and unified routing.}
Reader development uses 1,024 training and 256 observed development questions. Original generation token IDs are replayed through frozen Qwen3.5-4B using BF16 Hugging Face SDPA and FP32 head parameters and logits. Heads share labels and checkpoints at 0, 128, 256, 512, and 1,024 tokens. Mean development negative log-likelihood selects checkpoints under the three-seed, two-learning-rate search in Appendix~\ref{sec:hyperparameters}.

Development routing uses the same seed-29 answer-position head at 0 and 1,024 tokens. A fixed hash orders material groups within task, assigning 768 training groups to gate fitting, 128 to probability calibration, and 128 to threshold selection. Targets use out-of-fold reader predictions. The 256 development questions evaluate the complete-data reader and also inform reader selection.

\paragraph{Stopping and sentence scheduling.}
The fixed-grid pilot fixes reader seed 29 and uses four material-group folds for training readouts. All development questions enter reasoning, with checks at 128, 256, and 512 tokens and the terminal point. Sentence scheduling reuses natural 1,024-token trajectories and trains triggers on training prefixes, fixing entropy certainty at 0.85 and trigger probability at 0.25. Replay policies share candidate readouts obtained on an A100.

Joint replay crosses five initial gates and ten routing thresholds with terminal, every-eight-sentence, gap-4, and tick-4 schedules on aligned development traces. Each question incurs an initial read and follows its schedule if routed. These 200 configurations examine routing--stopping interactions on the observed development set.

\paragraph{Cost and statistical reporting.}
Replay reports reasoning calls, reads, triggers, and \emph{ideal generation tokens}, assuming instantaneous decisions once a prefix becomes observable. Online output additionally includes generation while reads finish and auxiliary query tokens.

The native-component evaluation measures GPU-synchronized pipeline time including generation, reads, branching, decisions, and exit. Inputs are pre-encoded; loading and between-trial bookkeeping are excluded. Pointwise 95\% intervals use 10,000 paired material-group bootstrap resamples stratified by the four ProofWriter subsets and LogiQA, with no noninferiority margin. The primary evaluation protocol appears in Section~\ref{sec:experiments}.

\section{Component Studies}
\label{sec:component-evidence}
\subsection{Measured Component Trade-offs}
\begin{table*}[!tbp]
\centering
\small
\setlength{\tabcolsep}{4pt}
\begin{tabular*}{\textwidth}{@{}@{\extracolsep{\fill}}lrrrrr@{}}
\toprule
Policy & Acc.\ (\%) & PW & LQ & Time (s) & Output tok. \\
\midrule
Native read, no reasoning & 49.74 & 115 & 267 & 0.0805 & 1.00 \\
Linear read, no reasoning & 65.76 & 252 & 253 & 0.0803 & 1.00 \\
Native read, always reason & 76.56 & 309 & 279 & 9.4488 & 1024.68 \\
Linear read, always reason & 79.04 & 333 & 274 & 9.4440 & 1024.68 \\
Linear, initial confidence 0.8 & 78.78 & 335 & 270 & 6.6186 & 710.80 \\
Linear, confidence 0.8 at all checks & 78.26 & 332 & 269 & 5.9578 & 613.52 \\
Linear, initial 0.8 / later 0.95 & 78.78 & 334 & 271 & 6.8394 & 706.76 \\
Linear, segmented; initial 0.8 only & 78.78 & 334 & 271 & 6.8943 & 713.10 \\
Linear, segmented; never stop early & 78.91 & 332 & 274 & 9.9169 & 1028.86 \\
Native read, long reasoning (8,192) & 84.64 & 355 & 295 & 43.9222 & 4704.13 \\
\bottomrule
\end{tabular*}
\caption{Frozen native-backbone component comparisons on 768 reserved questions. All reasoning budgets are 1,024 tokens except the long baseline. PW and LQ give correct counts out of 384 each. Seconds and output tokens are means from actual executions; output tokens include auxiliary reading requests. Segmented controls retain the fixed-grid reading and continuation procedure. These configurations precede the sentence-trigger experiments.}
\label{tab:native-confirmation}
\end{table*}

On the separate native-component cohort, the trained linear reader raises correctness from 588 to 607/768 at a 1,024-token budget and approximately 9.44 seconds per question (Table~\ref{tab:native-confirmation}). It corrects 44 answers and introduces 25 errors, gaining 2.47 percentage points (95\% CI: $[0.39,4.56]$). ProofWriter gains 24 correct answers while LogiQA loses five.

Initial confidence routing at 0.8 answers 236 questions directly and sends 532 to reasoning, yielding 605 correct in 6.6186 seconds. Relative to always reasoning with the linear reader, time decreases by 29.92\% (95\% CI: $[27.07,32.74]$), while accuracy changes by $-0.26$ points ($[-1.17,0.65]$).

Requiring 0.95 confidence at intermediate checks also yields 605 correct, but takes 6.8394 seconds; nine of the 532 reasoning trajectories stop early. It preserves every prediction from the segmented initial-only control and saves 0.80\% of time (95\% CI: $[0.27,1.41]$), yet runs 3.33\% slower than unsegmented initial routing. Lowering all thresholds to 0.8 yields 601 correct in 5.9578 seconds. Usable exit opportunities and checking overhead jointly determine the trade-off.

The long native baseline reaches 650/768 in 43.9222 seconds, with 173 traces truncated at 8,192 tokens. Its endpoint adds an answer read; parsing generated final answers instead gives 529 correct and 174 invalid formats. Longer reasoning reaches higher accuracy at substantially greater cost.

\subsection{Allocating Reasoning}
\begin{table*}[!tbp]
\centering
\small
\begin{tabular*}{\textwidth}{@{}@{\extracolsep{\fill}}lrrrrr@{}}
\toprule
Policy & Correct / 512 & Reasoning calls & Generated tokens & Corrected & Regressed \\
\midrule
Direct Kev & 311 & 0 & 0 & 0 & 0 \\
Native generation, parsed answer & 443 & 512 & 2,416,895 & --- & --- \\
Always reason, then Kev & 461 & 512 & 2,416,895 & 160 & 10 \\
Confidence routing & 450 & 364 & 1,840,852 & 149 & 10 \\
Margin routing & 452 & 371 & 1,863,812 & 151 & 10 \\
Entropy routing & 453 & 368 & 1,853,026 & 152 & 10 \\
Hidden initial-error gate & 452 & 379 & 1,884,419 & 151 & 10 \\
Hidden benefit gate & 455 & 367 & 1,743,003 & 150 & 6 \\
Four-feature benefit gate & 456 & 373 & 1,843,623 & 155 & 10 \\
Four-feature + hidden benefit gate & 456 & 369 & 1,752,851 & 151 & 6 \\
Always on PW; confidence on LQ & 457 & 376 & 1,867,037 & 156 & 10 \\
\midrule
Endpoint oracle (uses labels) & 471 & 160 & 768,115 & 160 & 0 \\
\bottomrule
\end{tabular*}
\caption{Frozen routing on 512 new questions, balanced across tasks. Gates use the 75\% calibration threshold; actual calls vary. Tokens come from paired trajectory replay capped at 8,192 per question. Corrected and regressed count changes relative to direct Kev. Parsed answers reuse the same native trajectories; 22 answers fail parsing.}
\label{tab:routing}
\end{table*}

The Kev study uses an external generator capped at 8,192 tokens. Reasoning followed by Kev yields 461/512 correct, compared with 311 for direct Kev, comprising 160 corrections and ten regressions (Table~\ref{tab:routing}). These two events define the benefit target.

At the 75\% calibration setting, the hidden benefit gate invokes reasoning 367 times and answers 455 correctly: 28.3\% fewer calls and 27.9\% fewer generated tokens than always reasoning, with six fewer correct answers. It captures 150/160 corrections and incurs six regressions. Against confidence routing, it gains five correct answers and saves 97,849 tokens; the paired accuracy-difference interval spans zero (95\% CI: $[-0.78,2.73]$ points).

Correction AUROC varies by task: benefit obtains 0.762 on ProofWriter and 0.668 on LogiQA, versus confidence's 0.597 and 0.819. Of the benefit gate's 367 calls, 211 leave endpoint correctness unchanged and consume 56.8\% of generated tokens.

\paragraph{Aligning the native endpoints.}
\begin{table*}[!tbp]
\centering
\small
\begin{tabular*}{\textwidth}{@{}@{\extracolsep{\fill}}lrrrr@{}}
\toprule
Gate & Correct/256 & Calls & Corrected & Regressed \\
\midrule
Hidden + four-feature benefit & 200 & 209 & 48 & 12 \\
Initial-error prediction & 201 & 185 & 49 & 12 \\
Confidence & 203 & 195 & 51 & 12 \\
Margin & 203 & 194 & 51 & 12 \\
Entropy & 203 & 197 & 51 & 12 \\
Benefit, global calibration & 195 & 193 & 41 & 10 \\
Benefit, task calibration & 202 & 201 & 46 & 8 \\
Task mean benefit & 193 & 197 & 40 & 11 \\
\bottomrule
\end{tabular*}
\caption{Unified native-reader routing on 256 observed development questions. All gates use the 75\% threshold-calibration setting; realized calls differ. Direct reading gives 164 correct answers, always reasoning gives 203, and the label-based endpoint oracle gives 217. Calls count reasoning invocations; each replay uses a 1,024-token cap. Corrected and regressed are relative to direct reading.}
\label{tab:unified-routing}
\end{table*}

The native-reader development comparison uses the same reader at both endpoints, yielding 164/256 direct and 203 terminal successes, with a label-based selection upper bound of 217 (Table~\ref{tab:unified-routing}). At the 75\% calibration setting, margin reaches 203 correct with 194 reasoning calls; benefit using the hidden state and four features reaches 200 with 209 calls. Global and task-conditional probability calibration reach 195 and 202. Benefit modeling does not improve on margin in this configuration.

\subsection{Extracting Answers from Prefixes}
\begin{table*}[!tbp]
\centering
\small
\setlength{\tabcolsep}{3pt}
\begin{tabular*}{\textwidth}{@{}@{\extracolsep{\fill}}lrrrrr@{}}
\toprule
Reader & 0 & 128 & 256 & 512 & 1,024 \\
\midrule
Answer position & $65.10\pm0.98$ & $65.89\pm1.13$ & $64.45\pm0.68$ & $70.18\pm1.93$ & $78.78\pm0.45$ \\
Natural final state & $39.84\pm2.07$ & $30.73\pm2.54$ & $30.60\pm1.58$ & $38.54\pm2.93$ & $41.15\pm2.15$ \\
Natural mean state & $33.20\pm4.88$ & $34.24\pm5.69$ & $33.07\pm4.95$ & $36.07\pm5.56$ & $45.18\pm4.82$ \\
Natural attention pool & $35.68\pm5.88$ & $38.41\pm7.01$ & $38.41\pm7.38$ & $42.71\pm7.89$ & $51.82\pm6.33$ \\
\bottomrule
\end{tabular*}
\caption{Matched heads on a frozen native backbone at the indicated reasoning-token budgets. The answer-position, final-state, and mean-state heads each have 10,240 parameters; attention pooling has 12,800. Development accuracy is reported in percent as mean $\pm$ standard deviation across three seeds. The 256 development questions also select checkpoints. Answer-position reading adds a closure and cue; natural-state interfaces use existing prefix states. Measurements use teacher-forced replay.}
\label{tab:readers}
\end{table*}

At 1,024 tokens, answer-position reading reaches 78.78\% mean development accuracy across three seeds, versus 41.15\%, 45.18\%, and 51.82\% for natural final-state, mean-state, and attention-pool heads (Table~\ref{tab:readers}). Heads share labels, prefix budgets, and optimization search; answer-position reading additionally processes a closure and answer cue.

This extraction gap motivates using natural states to select reading opportunities while retaining explicit answer queries.

\subsection{Reading Frequency and Stopping}
\begin{table}[!tbp]
\centering
\small
\setlength{\tabcolsep}{3pt}
\begin{tabular*}{\columnwidth}{@{}@{\extracolsep{\fill}}lrrr@{}}
\toprule
Schedule & Tokens & Reads & Triggers \\
\midrule
Terminal only & 1024.0 & 1.000 & 0.00 \\
Every 4 sentences & 788.2 & 10.719 & 0.00 \\
Every 8 sentences & 830.9 & 5.762 & 0.00 \\
Trigger every sentence & 779.7 & 6.363 & 41.31 \\
Trigger, gap-4 & 813.5 & 3.594 & 32.66 \\
Trigger, tick-4 & 855.0 & 2.570 & 11.07 \\
Trigger, tick-8 & 888.6 & 1.770 & 5.58 \\
\bottomrule
\end{tabular*}
\caption{Sentence scheduling on 256 observed development questions, with the same reader and entropy stopping threshold. All schedules yield 203/256 correct answers; tokens, reads, and triggers are per-question means. Token counts assume instantaneous decisions; this is trajectory replay, not an online timing comparison. Each early-stopping row loses one terminal success and recovers one terminal error.}
\label{tab:sentence-schedules}
\end{table}

All displayed sentence schedules yield 203/256 correct (Table~\ref{tab:sentence-schedules}). Each early-stopping schedule uses the same entropy threshold, recovering one terminal error and losing one terminal success. Reading every four sentences uses 788.2 ideal generation tokens and 10.719 reads per question; every eight uses 830.9 tokens and 5.762 reads.

A learned gap-4 trigger uses 813.5 tokens, 3.594 reads, and 32.66 trigger calls. Tick-8 reduces reads to 1.770 and triggers to 5.58 while delaying stopping to 888.6 ideal tokens. Sparser observation exchanges fewer queries for longer generation.

\begin{table}[!tbp]
\centering
\small
\setlength{\tabcolsep}{3pt}
\begin{tabular*}{\columnwidth}{@{}@{\extracolsep{\fill}}lrrrr@{}}
\toprule
Rule & $\tau$ & Correct & Reads & Tokens \\
\midrule
Terminal only & --- & 203 & 1.00 & 1,024.0 \\
Raw confidence & .90 & 198 & 3.04 & 703.0 \\
Raw confidence & .95 & 202 & 3.50 & 852.0 \\
Learned correctness & .90 & 199 & 3.73 & 928.0 \\
Learned correctness & .95 & 203 & 3.98 & 1,018.5 \\
Joint, $\lambda=0$ & .90 & 203 & 3.98 & 1,017.0 \\
Joint, $\lambda=.05$ & .90 & 199 & 3.69 & 919.5 \\
Joint, $\lambda=.05$ & .95 & 202 & 3.96 & 1,009.0 \\
\bottomrule
\end{tabular*}
\caption{Stopping pilot on 256 observed development questions with a fixed answer-position reader. All questions enter reasoning. Tokens assume instantaneous decisions; reads count performed calls. Displayed settings were specified before stop-model fitting.}
\label{tab:stopping}
\end{table}

In the fixed-grid pilot, confidence at 0.95 yields 202/256 correct, 852 ideal tokens, and 3.50 reads (Table~\ref{tab:stopping}). The joint correctness--continuation rule at the same threshold and $\lambda=0.05$ also yields 202, but uses 1,009 tokens and 3.96 reads. Checking confidence only at 128 and 512 preserves the dense grid's predictions with 863 tokens and 2.656 reads, improving observation cost over the tested joint target.

\paragraph{Interaction with initial routing.}
On the same development traces, margin routing sends 194 questions to reasoning and yields 203/256 correct. Every-eight-sentence stopping reduces mean ideal generation from 776.0 to 694.3 tokens per input. Gap-4 uses 691.9 tokens and lowers total reads from 5.891 to 3.996, adding 28.32 triggers; tick-4 uses 702.8 tokens, 3.078 reads, and 9.22 triggers. Reads include the initial query, and all three schedules preserve routed questions' terminal correctness.

\subsection{Online Component Execution}
The online development study compares terminal reading, every-eight-sentence reading, and gap-4, tick-4, and tick-8 triggers on 32 observed questions with three repetitions. All questions enter reasoning; policies run in shuffled order on the same A100 for each question, with fresh generation per policy. This eager vLLM configuration enables cross-request prefix caching, unlike the compiled, cache-disabled primary evaluation.

Terminal reading yields 22/32 correct in 60.593 seconds; every-eight-sentence reading yields 23 in 48.651 seconds, gap-4 yields 25 in 48.351 seconds, and tick-4 yields 23 in 47.839 seconds. The three intermediate-reading policies change generated prefixes on 24, 16, and 13 questions, respectively. These are complete-policy comparisons with trajectory changes.

\end{document}